\documentclass[preprint,review,12pt]{elsarticle}
\journal{International Journal of Artificial Intelligence in Education}

\usepackage[utf8]{inputenc}
\usepackage[T1]{fontenc}
\usepackage{hyperref}
\usepackage{doi}
\usepackage{booktabs}
\usepackage{graphicx}
\usepackage{amsmath, amssymb}

\hypersetup{colorlinks=true, linkcolor=blue, citecolor=blue, urlcolor=blue, pdfauthor={}}

\usepackage{cleveref}

\begin{document}

\begin{frontmatter}
\title{Making Reasoning Visible: Configuration-Aware Diagnostic Modeling for Adaptive Scaffolding in Structured Reasoning Tasks}
\author{
  Trystan Geoffre\textsuperscript{1,*}\,,
  Zheng Yuan\textsuperscript{2}\,,
  Thierry Geoffre\textsuperscript{3}
  \\[2ex]
  \small
  \textsuperscript{1}University of Fribourg, Switzerland\\
  \textsuperscript{2}University of Sheffield, England\\
  \textsuperscript{3}University of Luxembourg, Luxembourg
  \\[2ex]
  \small
  \textsuperscript{*}Corresponding author: \href{mailto:trystan.geoffre@gmail.com}{trystan.geoffre@gmail.com}
  \\[1ex]
  \small
  Emails:
  \href{mailto:trystan.geoffre@gmail.com}{trystan.geoffre@gmail.com},
  \href{mailto:zheng.yuan1@sheffield.ac.uk}{zheng.yuan1@sheffield.ac.uk},
  \href{mailto:thierry.geoffre@uni.lu}{thierry.geoffre@uni.lu}
}

\begin{abstract}
Modeling the latent, multi-step reasoning processes of learners is a core challenge for adaptive tutoring systems, which often rely on simple right/wrong answers rather than understanding \emph{how} students think. Instantiated here for grammatical reasoning, this study introduces a diagnostic framework that makes students' step-by-step problem-solving visible by analyzing their action sequences in an interactive learning game. The framework computes five interpretable strategy dimensions (e.g., systematic error correction, random guessing), estimates coordination demand from task structure and behavioral patterns, and classifies reasoning into four distinct states. Analyzing 36,234 interactions from 234 students working on 42 morphosyntactic exercises, we found that reasoning strategies vary primarily by context rather than stable traits (intraclass correlation = 0.14), and students show high persistence in unproductive states; specifically, 74.6\% of failures driven by high coordination demand (rather than simple knowledge gaps) are followed by another failure without intervention. Our early warning model predicts validation success with strong accuracy (AUC = 0.826 using strategy patterns alone; 0.902 with additional sequential context) while avoiding circular reliance on current error counts. These findings demonstrate that making latent reasoning processes visible enables earlier, more targeted scaffolding that responds to \emph{how} students are thinking rather than simply whether they succeed. While demonstrated in language learning, the framework provides a generalizable, interpretable architecture for diagnosing complex, structured reasoning in any domain.
\end{abstract}

\end{frontmatter}

\section{Introduction}
Intelligent tutoring systems (ITS) have fundamentally transformed educational technology by providing personalized, step-by-step guidance. Grounded in cognitive architectures like Adaptive Control of Thought—Rational (ACT-R) and formalized through inner-loop (step-level feedback) and outer-loop (task selection) designs, these systems excel in well-defined procedural domains where errors map directly to missing knowledge components \cite{anderson1995cognitive, vanlehn2006behavior}. In parallel, recent advances in learning analytics and stealth assessment have begun to infer latent competencies from clickstream data, demonstrating that action sequences, timing, and choice patterns can reveal underlying reasoning strategies \cite{shute2011stealth}. Together, these approaches have established a robust foundation for tracking \emph{what} a student knows and \emph{whether} they can apply it correctly.

However, these frameworks struggle when applied to complex, non-linear reasoning tasks like morphosyntax. Recent sequence-based learning analytics have revealed that, in this domain, learners do not simply retrieve static rules; rather, they treat every individual action as a dynamic hypothesis test about linguistic structure, converging toward a solution through highly individualized paths \cite{geoffre2026modeling}. Because language is a fundamentally nonalgorithmic domain—where the inherent irregularity of spelling or morphosyntax means no single strategy guarantees a correct outcome \cite{rittlejohnson1999learning}—students are forced to constantly navigate multiple competing hypotheses, consistent with Siegler's overlapping waves theory \cite{siegler2007cognitive}. In this context, learners are juggling probabilistic, uncertain approaches rather than executing deterministic production rules. Consequently, traditional ITS models cannot reliably map these fluid actions to static knowledge states, and standard outcome-based metrics fail to capture the rich reasoning dynamics that precede a final answer. Scaffolding systems are thus left blind to whether a student lacks foundational knowledge or is simply overwhelmed by the coordination demands of the task, risking interventions that are misaligned with the student's actual thought process and disruptive to productive struggle \cite{aleven2017instruction}.

To address this gap, we introduce a theory-driven diagnostic framework that makes hidden reasoning processes observable and actionable in real time. Rather than relying on opaque black-box predictors that lack pedagogical transparency \cite{roll2016evolution} or retrospective outcome metrics, our pipeline transforms raw interaction sequences into interpretable, pedagogically meaningful signals. Specifically, the framework concurrently computes five strategy dimensions (e.g., systematic error correction, random guessing) and estimates a Coordination Demand Index that captures both structural task complexity and behavioral struggle. These signals are integrated into a transparent, state-aware classification system that maps each reasoning cycle to distinct cognitive states. By explicitly excluding direct error counts from its predictive logic to avoid circularity, the framework provides a non-circular early warning mechanism. This enables a shift from uniform, outcome-based feedback to real-time truly adaptive scaffolding that responds to \emph{how} a student is reasoning, preserving learner autonomy during stable states while triggering targeted support during unproductive persistence.

We validate this framework through a large-scale analysis of 36,234 interactions from 234 students across 42 distinct morphosyntactic exercises. Our empirical findings demonstrate that reasoning strategies are fundamentally state-dominant rather than trait-like (mean intraclass correlation = 0.14), justifying the need for real-time, context-aware adaptation over static profiling. Furthermore, we reveal that students exhibit high persistence in unproductive states---specifically, 74.6\% of failures driven by high coordination demand are followed by another failure without intervention. Our early warning model predicts validation success with strong accuracy (AUC = 0.826 using strategy patterns alone; 0.902 with sequential context), substantially outperforming conventional ITS baselines. Ultimately, while instantiated here for language learning, this study provides a generalizable, interpretable architecture for diagnosing complex, structured reasoning. The remainder of this paper is structured as follows: Section 2 reviews related work; Section 3 details the framework; Section 4 presents empirical results; Section 5 discusses theoretical implications and scaffolding design; and Section 6 concludes.

\section{Related Work}

\subsection{Intelligent Tutoring Systems and Adaptive Learning Foundations}
Intelligent tutoring systems have long relied on structured architectures to diagnose learner states and deliver targeted support. Model-tracing approaches rooted in cognitive architectures like ACT-R pioneered step-level diagnosis, demonstrating how systems can map discrete student actions to specific knowledge components for precise error remediation \cite{anderson1995cognitive}. Expanding on this foundation, VanLehn's \cite{vanlehn2006behavior} influential framework formalized the distinction between the inner loop (step-level feedback) and outer loop (task sequencing), establishing a comprehensive blueprint for real-time pedagogical decision-making. Meta-analytic evidence confirms that intelligent tutoring systems can substantially improve learning outcomes compared to traditional instruction, particularly in well-defined domains \cite{kulik2016effectiveness}. More recently, adaptive learning technologies have expanded beyond knowledge tracing to incorporate pedagogical design principles that tailor instruction to individual learner trajectories \cite{aleven2017instruction}. While highly effective for procedural and well-structured domains, these frameworks primarily track the accumulation, application, or decay of discrete knowledge components. They are less equipped to handle reasoning processes characterized by non-linear hypothesis testing, competing cognitive demands, and rapid strategy switching. In morphosyntactic tasks, learners frequently revise partial solutions, backtrack across agreement chains, and balance accuracy against working memory constraints. Capturing these fluid cognitive dynamics requires moving beyond static component tracing toward a diagnostic architecture that continuously evaluates strategy configurations and coordination demand in real time.

\subsection{Feedback Theory and Formative Assessment}
Effective scaffolding is fundamentally grounded in feedback theory and formative assessment practices. Hattie and Timperley's \cite{hattie2007power} meta-analytic framework establishes that high-impact feedback must address three questions: Where am I going? How am I going? Where to next? Similarly, Shute \cite{shute2008focus} distinguishes between directive feedback (providing explicit corrections) and facilitative feedback (prompting self-regulation), emphasizing that the latter better supports long-term retention and transfer. Sadler's \cite{sadler1989formative} seminal work further argues that formative assessment is only effective when learners can perceive the gap between current and target performance and possess actionable pathways to close it. Empirical studies in online tutoring confirm that immediate feedback during problem solving improves learning outcomes \cite{kehrer2013does}. Yet, these models implicitly assume that the performance gap is either observable or easily articulated. In grammatical reasoning, the gap is often hidden behind unproductive strategy configurations, elevated cognitive load from managing multiple interdependent constraints, or flawed hypothesis testing (formalized as the Coordination Demand Index in Section~\ref{sec:ccl}). Students may possess the requisite grammatical knowledge but fail to apply it due to working memory constraints or misdirected attention. Our framework addresses this by diagnosing the gap in real time from process telemetry, transforming latent reasoning states into quantifiable signals that trigger context-aware, facilitative interventions precisely when the gap becomes pedagogically actionable.

\subsection{Stealth Assessment and Configuration-Aware Diagnosis}
The rise of game-based and digital learning environments has catalyzed stealth assessment methodologies, which infer learner competencies from in-game behaviors without disrupting the learning flow \cite{shute2011stealth, shute2013stealth}. By mapping clickstreams, response times, and choice patterns to latent traits, researchers have successfully quantified agency \cite{snow2015agency}, modeled probabilistic relationships between actions and competencies using Bayesian networks \cite{lu2025from}, and tracked learning curves to identify knowledge acquisition bottlenecks \cite{harpstead2015using}. Process mining techniques have further enabled the discovery of sequential behavioral patterns that predict learning styles and strategy preferences \cite{fatahi2018empirical, bannert2014process, bechet2012discovering}, while educational data mining reviews have highlighted the importance of distinguishing productive struggle from unproductive wheel-spinning \cite{bakeryacef2009state}. Despite these advances, most stealth assessment approaches aggregate performance over extended periods or infer static competency profiles. They rarely capture the moment-to-moment interplay of multiple reasoning strategies or the dynamic shifts in coordination demand that precede validation outcomes. Our framework advances this tradition by modeling strategy configurations as concurrent, continuously updated signals. Rather than inferring whether a student ``knows'' a grammatical rule, we diagnose how they are reasoning about it in real time, enabling scaffolding that responds to transient cognitive states rather than retrospective proficiency estimates.

Nevertheless, most current detectors of unproductive persistence  rely on coarse behavioral aggregates and black‑box models.  For instance, Liu et al.\ \cite{liu2023struggling} trained a struggle  detector for an open‑ended science game, achieving AUC = 0.635,  insufficient for individual intervention.  Karumbaiah et al.\ \cite{karumbaiah2018predicting} predicted level  quitting in a physics game with AUC = 0.81 using gradient boosting  and 101 features aggregated over 60‑second windows, but the model  required post‑hoc partial dependence plots for interpretation.  Within grammar ITSs, Deininger et al.\ \cite{deininger2023firsttry}  predicted first‑attempt correctness on individual exercise fields  in an English tutoring system. Their best model (XGBoost) reached  AUC = 0.68, while logistic regression yielded only 0.57; the  features were limited to prior performance, time, and attempt  count, without any representation of reasoning strategies. These findings indicate that even in structured domains,  standard behavioral features leave considerable predictive power untapped, and that transparent models often lag behind complex  ensembles. Our framework directly addresses this gap by replacing  opaque outcome‑based features with five theory‑driven strategy  dimensions and a coordination demand index, all computed in real  time, and by showing that an inherently interpretable logistic  regression can outperform the black‑box alternatives.

\subsection{Metacognition, Self-Regulated Learning, and Cognitive Variability}
Metacognitive awareness and self-regulated learning (SRL) are critical drivers of complex reasoning performance. Flavell's \cite{flavell1979metacognition} foundational work established metacognition as the active monitoring and regulation of cognitive processes, while Winne and Hadwin's \cite{winne1998studying} COPES model operationalized SRL as a cyclical process of condition evaluation, strategy selection, and outcome monitoring. A large-scale meta-analysis has confirmed a consistent, positive relationship between self-regulated learning and academic achievement across childhood and adolescence \cite{dent2016relation}. In grammatical reasoning, these processes manifest as learners' ability to detect mismatches, adjust chain-completion priorities, and abandon unproductive hypotheses. However, traditional SRL measurement relies heavily on self-report inventories or think-aloud protocols, which are intrusive, retrospective, and unable to capture rapid strategy shifts during task execution. Moreover, Siegler's \cite{siegler2007cognitive} overlapping waves theory demonstrates that cognitive variability is not noise but a signature of learning: students simultaneously hold multiple strategies and switch between them based on perceived efficacy and task demands. Our framework operationalizes metacognitive regulation directly from action sequences, translating theoretical SRL constructs into observable, validation-level metrics. By quantifying strategy stability, error-targeting precision, and adaptive displacement, we provide a real-time lens into how learners monitor and adjust their reasoning without interrupting the natural flow of practice.

\subsection{Interpretable Artificial Intelligence for Educational Decision-Making}
As machine learning becomes increasingly embedded in educational technologies, the tension between predictive accuracy and pedagogical transparency has intensified. Rudin \cite{rudin2019stop} argues forcefully against post-hoc explanations of black-box models in high-stakes domains, advocating instead for inherently interpretable algorithms whose decision logic can be audited and trusted by practitioners. In education, this transparency is essential: teachers and system designers must understand why a scaffolding trigger was activated to align interventions with pedagogical goals and avoid hint fatigue \cite{holstein2019codesigning, dellermann2019hybrid}. Surveys of interpretability methods emphasize that feature importance, coefficient transparency, and decision boundary clarity are prerequisites for educational deployment \cite{molnar2020interpretable, carvalho2019machine}. Yet, many learning analytics pipelines still prioritize complex ensembles or deep architectures at the expense of explainability. Our framework explicitly adopts interpretable logistic regression with constrained feature engineering, ensuring that every prediction is traceable to specific strategy scores and coordination demand estimates. This design choice prioritizes pedagogical actionability over marginal gains in predictive performance, enabling educators to audit, adapt, and trust the scaffolding logic.

\subsection{Grammatical Reasoning and Intelligent Language Tutoring}
The acquisition of morphosyntax is governed by frequency effects, deliberate practice, and the gradual automation of rule application \cite{ellis2002frequency, dekeyser2007practice}. Intelligent language tutoring systems (ILTS) have historically relied on natural language parsers to analyze open-ended learner input, diagnosing errors in agreement, tense, and reference with high granularity \cite{heift2007errors, karmiloffsmith1981functional}. While parser-based approaches yield rich linguistic diagnostics, they are computationally intensive, domain-specific, and difficult to scale across diverse classroom contexts. Recent curriculum analyses in French education further emphasize that morphosyntactic agreement remains a high-stakes bottleneck, requiring targeted intervention that moves beyond binary correctness scoring \cite{brissaud2018etude, ravid2002developing}. Recent work in English grammar ITSs further supports a  configuration‑aware perspective: Deininger et al.\  \cite{deininger2025who} demonstrated that the same learning  behaviors (e.g., corrections after feedback) were positively  associated with achievement for some learners and negatively  for others, depending on their prior knowledge and motivation.  This underscores the risk of static, aggregate behavioral  profiles and the need for real‑time diagnosis that accounts  for the learner's current cognitive and strategic state. Moreover, studies of automated corrective feedback reveal that learner uptake depends heavily on cognitive readiness and motivational state, not merely error detection \cite{ranalli2018automated}. Our framework bypasses the scalability limitations of parser-based ILTS by leveraging structured game telemetry to extract grammatical reasoning patterns at scale. By quantifying chain competition, error-targeting efficiency, and validation pressure, we provide a computationally lightweight yet pedagogically precise alternative that aligns with frequency-based acquisition models and contemporary language curriculum demands.

\section{Diagnostic Framework \& Methodology}

This section details the diagnostic framework that transforms raw interaction logs into interpretable reasoning states for adaptive scaffolding. We first provide a high-level overview of the pipeline, then describe the learning environment and dataset, followed by detailed specifications of each computational component.

\subsection{Framework Overview}
\label{sec:framework_overview}

\begin{figure}[ht]
    \centering
    \includegraphics[width=\textwidth]{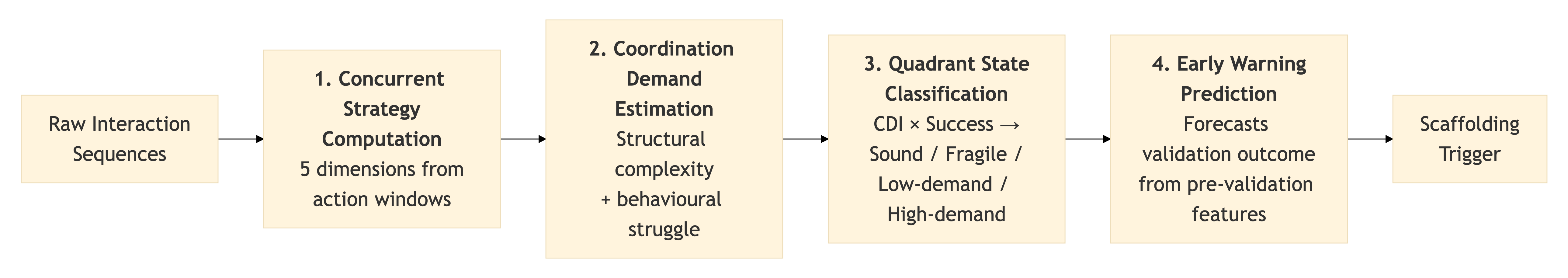}
    \caption{High-level architecture of the diagnostic framework. Raw interaction sequences (slider movements and validations) are processed through four integrated components to produce interpretable reasoning states and early warning predictions that inform adaptive scaffolding decisions.}
    \label{fig:diagnostic_pipeline}
\end{figure}

As illustrated in Figure~\ref{fig:diagnostic_pipeline}, the diagnostic pipeline operates in real time on each validation event, transforming low-level interaction sequences into actionable pedagogical signals through four integrated components:

\begin{enumerate}
    \item \textbf{Concurrent Strategy Computation} (Section~\ref{sec:strategies}): Five interpretable dimensions quantify distinct reasoning patterns from action windows between validations.
    \item \textbf{Coordination Demand Estimation} (Section~\ref{sec:ccl}): A composite index combines structural task complexity with behavioral struggle indicators.
    \item \textbf{Quadrant-Based State Classification} (Section~\ref{sec:quadrants_transitions}): Cross-referencing demand and success yields four distinct reasoning states that distinguish knowledge gaps from coordination bottlenecks.
    \item \textbf{Early Warning Prediction} (Section~\ref{sec:prediction}): An interpretable logistic regression model forecasts validation success using pre-validation features alone, enabling proactive scaffolding triggers.
\end{enumerate}

\noindent The following subsections detail the learning environment, dataset characteristics, and mathematical specifications for each component. Full equations, edge-case handling, and computational logic are provided in Appendix A; comprehensive robustness checks and generalization analyses are documented in Appendix B.

\subsection{Learning Environment \& Data Collection}
\label{sec:data_collection}
Data were collected from an interactive grammatical reasoning game deployed on the publicly funded, non-commercial platform [Reference blinded for review], developed in partnership with cantonal educational authorities in French-speaking Switzerland. The game asks players to build grammatically correct sentences by moving sliders, each containing several words or different inflections of the same word. By repositioning sliders, players align words to respect agreement chains (e.g., subject-verb, determiner-noun, adjective-noun) and produce syntactically valid sentences. 

When a player believes a sentence may be correct, they can submit it for validation and receive binary feedback (correct/incorrect). The game deliberately withholds explicit information about underlying agreement chains and does not localize errors (e.g., indicating which sliders are incorrect). By providing only binary outcomes, the design compels students to observe individual slots and infer syntactic dependencies through systematic trial, reflection, and hypothesis testing. This approach intentionally mirrors authentic language acquisition, where morphosyntactic rules are internalized through pattern recognition rather than explicit instruction, and we have consistently observed progressively deeper attentional engagement with this gameplay mechanic.

Data were collected over a one-year period (April 2025--March 2026) from 11 classes across 8 schools in French-speaking Switzerland with primary school students aged 8-11, where the game was freely available for classroom use alongside other platform activities. The research was conducted with full informed consent from parents and teachers, in compliance with all applicable ethical guidelines; a complete ethics statement and study procedure description will appear in the final publication and are omitted here to preserve double-blind anonymity. This particular game was selected for analysis due to its fine-grained decomposition of morphosyntactic reasoning task, yielding high-resolution action sequences suitable for diagnostic modeling.

Each logged interaction includes: anonymized student identifier, class identifier, session start/end timestamps (enabling duration calculation), validation outcome (correct/incorrect), and a temporally ordered sequence of actions. Actions are of two primary types: \texttt{tirette} (slider movement, recording which slider was moved and to which position) and \texttt{validate} (submission attempt, recording the full sentence state at validation time). This telemetry enables reconstruction of the student's reasoning trajectory at the level of individual grammatical decisions.

\subsection{Dataset Composition \& Coverage}
\label{sec:dataset_composition}
The analytic dataset comprises 36,234 logged interactions from 234 students across 2,144 student-exercise sessions and 42 distinct morphosyntactic exercises. Validation events ($n=10,046$) serve as the primary diagnostic units, with tirette (drag) actions ($n=26,188$) providing the sequential substrate for strategy computation.

Exercise complexity varies by syntactic structure. Table~\ref{tab:sliders} shows the distribution of exercises by number of sliders (i.e., grammatical slots requiring alignment), ranging from 2 to 13. Table~\ref{tab:solutions} shows the distribution by number of valid solution patterns per exercise, ranging from 1 to 20. This variability ensures that diagnostic signals are evaluated across diverse cognitive demands and solution spaces.

\begin{table}[h]
\centering
\caption{Distribution of exercises per number of sliders}
\label{tab:sliders}
\begin{tabular}{cc}
\toprule
Number of sliders & Number of exercises \\
\midrule
2 & 3 \\
3 & 7 \\
4 & 2 \\
5 & 1 \\
6 & 2 \\
7 & 4 \\
8 & 9 \\
9 & 6 \\
10 & 4 \\
11 & 3 \\
12 & 1 \\
13 & 1 \\
\bottomrule
\end{tabular}
\end{table}

\begin{table}[h]
\centering
\caption{Distribution of exercises per number of solutions}
\label{tab:solutions}
\begin{tabular}{cc}
\toprule
Number of solutions & Number of exercises \\
\midrule
1 & 1 \\
2 & 6 \\
3 & 4 \\
4 & 7 \\
5 & 2 \\
6 & 8 \\
8 & 6 \\
9 & 1 \\
10 & 1 \\
12 & 1 \\
16 & 2 \\
18 & 3 \\
20 & 1 \\
\bottomrule
\end{tabular}
\end{table}

Validation success rate is 38.1\%, indicating moderate task difficulty appropriate for diagnostic sensitivity. The basic composition of the dataset is summarized in Table~\ref{tab:data_basic}.

\begin{table}[h]
\centering
\caption{Basic dataset composition}
\label{tab:data_basic}
\begin{tabular}{llr}
\toprule
\textbf{Component} & \textbf{Description} & \textbf{Count} \\
\midrule
Total events & All logged interactions & 36,234 \\
Unique students & Distinct learner identifiers & 234 \\
Total sessions & Student-exercise instances & 2,144 \\
Validation events & \texttt{validate} actions & 10,046 \\
Tirette actions & \texttt{tirette} (drag) actions & 26,188 \\
Exercises & Distinct task types & 42 \\
\midrule
\textbf{Validation Outcomes} \\
Success rate & $\text{validation\_success} = \text{True}$ & 3,828 (38.1\%) \\
Failure rate & $\text{validation\_success} = \text{False}$ & 6,218 (61.9\%) \\
\bottomrule
\end{tabular}
\end{table}

\subsection{Data Preprocessing \& Feature Engineering}
\label{sec:data_preprocessing}
Raw event streams are filtered to remove interface noise, incomplete interactions, and rapid-fire sequences indicative of wheel-spinning or gaming behaviors \cite{bakeryacef2009state}. Valid actions are mapped to state vectors $\mathbf{v}_t \in \{0,1\}^K$, where $K$ denotes the number of grammatical slots in the current exercise. Each state vector is compared against a set of gold solutions $\mathbf{g}^{(j)}$ to compute the Hamming distance \cite{hamming1950error} $h_t$, representing the minimum number of mismatches relative to any correct assignment pattern. Following Geoffre \& Geoffre \cite{geoffre2026modeling}, this distance metric is explicitly used to distinguish the closest candidate solution from non-optimal alternatives, enabling real-time tracking of hypothesis refinement. Sessions are temporally aligned, and validation events are extracted as discrete diagnostic windows.

\subsection{Concurrent Strategy Computation}
\label{sec:strategies}
A core contribution of this framework is the operationalization of five interpretable, concurrent strategy dimensions that transform opaque action sequences into quantifiable reasoning signals. Rather than treating grammatical reasoning as a black box or reducing it to binary success/failure, these dimensions were iteratively designed to bridge linguistic processing constraints, cognitive load theory, and metacognitive regulation models. This concurrent computation approach aligns with process-mining methodologies that extract pedagogically meaningful patterns from sequential educational data \cite{bannert2014process}, while extending linguistic pattern discovery techniques to grammatical reasoning contexts \cite{bechet2012discovering}. Each strategy captures a distinct pedagogical construct, is normalized to a [0,1] scale for direct interpretability in scaffolding decisions, and is computed in real time from validation-level action windows. Full mathematical specifications, edge-case handling, and normalization procedures are provided in Appendix A (Equations 1--20).

The five dimensions are defined as follows:

\begin{itemize}
    \item \textbf{Verb-First targeting} ($S_{\text{verb}}$) quantifies whether students prioritize the verb slot early in their reasoning cycle and maintain sustained attention on its agreement chain. Sequence-based analysis in morphosyntactic games reveals that learners frequently anchor on the verb before adjusting preceding elements, a non-linear processing pattern that aligns with pedagogical recommendations for identifying the syntactic pivot in French \cite{geoffre2026modeling}. This strategy reflects early functional category acquisition and efficient working memory allocation during agreement chain resolution \cite{karmiloffsmith1981functional}. The dimension rewards systematic anchoring rather than peripheral tinkering (Equations 1--6).
    
    \item \textbf{Chain-Aware progression} ($S_{\text{chain}}$) measures whether students complete or make measurable progress on one agreement chain before switching to another. Cognitive load theory posits that sequential dependency resolution minimizes extraneous load when working memory is taxed, whereas frequent context-switching fragments attention and inflates coordination demand \cite{sweller1994cognitive}. This dimension captures adherence to sequential resolution principles, operationalizing systematic hypothesis testing across morphosyntactic structures (Equations 7--10).
    
    \item \textbf{Error-Focus} ($S_{\text{focus}}$) computes the proportion of actions that actively reduce structural mismatch (Hamming distance) relative to the closest valid configuration. This operationalizes targeted error correction as a metacognitive monitoring strategy, directly linking to self-regulated learning frameworks where productive struggle is characterized by deliberate mismatch detection and feedback utilization rather than random adjustment \cite{winne1998studying, aleven2017instruction}. By quantifying the precision of error-targeting, this dimension distinguishes systematic correction from superficial trial-and-error (Equations 11--12).
    
    \item \textbf{Randomness} ($S_{\text{rand}}$) captures undirected exploration by combining chain-switching entropy with low error-reduction rates. While some exploration is necessary for hypothesis generation (Siegler's overlapping waves theory), excessive randomness indicates disorientation or unguided guessing. This dimension explicitly distinguishes productive hypothesis testing from counterproductive thrashing, enabling scaffolding systems to intervene when exploration ceases to yield structural progress \cite{siegler2007cognitive} (Equations 13--15).
    
    \item \textbf{Strategic Commitment} ($S_{\text{strat}}$) identifies sustained pursuit of a non-initial candidate solution until it becomes at least as close as the previously optimal path. This operationalizes hypothesis persistence, capturing whether learners commit to a structural hypothesis and systematically refine it rather than prematurely abandoning viable configurations. High commitment scores reflect resistance to impulsive switching and align with convergence patterns observed in expert-like problem-solving trajectories \cite{siegler2007cognitive} (Equations 16--18).
\end{itemize}

\noindent\textbf{Computation \& Real-Time Windowing}: Strategy scores are computed concurrently from action sequences within validation-level windows ($\mathcal{W}_v$), defined as all \texttt{tirette} actions occurring between two consecutive \texttt{validate} events. This windowing aligns with the natural reasoning cycle of hypothesis formulation, structural adjustment, and submission, enabling real-time diagnosis without retrospective aggregation. When a window contains fewer than two valid actions (e.g., immediate resubmission), the framework gracefully falls back to session-level aggregation to maintain continuous diagnostic coverage. 

All five dimensions are computed in parallel and normalized independently to account for exercise-specific structural constraints, ensuring consistent interpretability across diverse task formats. Their concurrent availability provides a multi-dimensional view of student reasoning, capturing distinct cognitive and metacognitive patterns within each diagnostic window. These strategy signals form the foundational input for subsequent cognitive demand estimation and reasoning state classification, detailed in the following sections, and ultimately inform the transparent scaffolding policy outlined in Section~\ref{sec:scaffolding_policy}. This strategy-mining pipeline demonstrates how fine-grained telemetry can yield pedagogically precise diagnostics at scale, extending prior work on behavioral pattern extraction \cite{bannert2014process, bechet2012discovering} to the domain of morphosyntactic reasoning.

 \subsection{Feature Coverage \& Data Quality}
\label{sec:feature_coverage}
Because the diagnostic framework relies on exercise-specific structural constraints, feature coverage varies by dimension. Table~\ref{tab:feature_coverage} summarizes the availability of each computed signal at the validation level. 

Strategy score coverage is constrained by the syntactic structure of the exercises. For instance, $S_{\text{verb}}^{\text{(val)}}$ is defined for 62.3\% of validations, as it requires the presence of a verb agreement chain in the current exercise. The remaining four strategy dimensions achieve 86.7--86.9\% coverage. Missing values reflect genuine exercise constraints (e.g., single-chain tasks preclude chain-aware computation) rather than pipeline failure or missing telemetry. Coordination demand estimates ($I_{\text{combined}}$) (detailed in Section~\ref{sec:ccl}) are computed for 100\% of actions via forward-fill alignment from preceding drag events. 

All downstream analyses (quadrant assignment, prediction modeling) use complete-case filtering based on the available features for a given exercise type; no imputation is applied. Strategy scores require $\geq 2$ valid \texttt{tirette} actions in the validation window; when fewer actions occur, the framework falls back to session-level aggregation as specified in Section~\ref{sec:strategies}.

\begin{table}[h]
\centering
\caption{Feature coverage and availability (validation-level)}
\label{tab:feature_coverage}
\begin{tabular}{llr}
\toprule
\textbf{Feature Category} & \textbf{Description} & \textbf{Coverage} \\
\midrule
\multicolumn{3}{l}{\textbf{Strategy Dimensions}} \\
$S_{\text{verb}}^{\text{(val)}}$ & Verb-First targeting & 6,257 (62.3\%) \\
$S_{\text{chain}}^{\text{(val)}}$ & Chain-Aware progression & 8,730 (86.9\%) \\
$S_{\text{rand}}^{\text{(val)}}$ & Randomness & 8,730 (86.9\%) \\
$S_{\text{focus}}^{\text{(val)}}$ & Error-Focus & 8,730 (86.9\%) \\
$S_{\text{strat}}^{\text{(val)}}$ & Strategic Commitment & 8,712 (86.7\%) \\
\midrule
\multicolumn{3}{l}{\textbf{Coordination Demand Index}} \\
$I_{\text{combined}}$ (tirette) & Structural + Behavioral & 26,188 (100\%) \\
$I_{\text{combined}}$ (validation) & Forward-filled from tirette & 10,046 (100\%) \\
\bottomrule
\end{tabular}
\end{table}

\subsection{Coordination Demand Index (CDI) Estimation}
\label{sec:ccl}
The Coordination Demand Index (CDI) quantifies the working memory demand imposed by competing agreement chains and observable struggle behaviors.\footnote{CDI serves as a computational proxy for task-structure and behavioral demand. It does not claim to measure subjective cognitive load as defined by psychometric frameworks (e.g., Sweller, 1988) or assessed via self-report scales (e.g., Paas, 1992).} The metric decomposes into structural and behavioral interference components, integrated via a theoretically grounded conditional combination rule.

\textbf{Structural interference} ($I_t^{\text{struct}}$) proxies the working-memory demands typically associated with competing chains. It weights the proportion of non-target chains by a normalized capacity factor and incorporates both average and maximum error distribution across those chains. A baseline coordination cost ensures non-zero demand whenever competing dependencies exist, even when perfectly matched (Equations 21--24).

\textbf{Behavioral interference} ($I_t^{\text{behav}}$) detects struggle patterns using a trailing event window ($w=3$). It aggregates four temporal and interaction signals: rapid guessing (burst frequency $<1.5$s), prolonged hesitation (pause threshold $\theta_{\text{pause}}$), slot cycling (revisiting identical slots), and validation pressure (excessive checking attempts) (Equations 25--29). The 1.5s burst threshold reflects the minimum inter-action time for deliberate manipulation, while $\theta_{\text{pause}}$ is calibrated to the 90th percentile of inter-event times to capture unusually long hesitations relative to individual pacing. Full calibration procedures, including clipping bounds and dataset-specific derivation logic, are documented in Appendix B, Section 2.1.

The \textbf{conditional combination} rule is grounded in cognitive load theory:  moderate-to-high intrinsic load, especially when element interactivity  is high, can impair learners' ability to monitor and regulate their own  strategies \cite{sweller2019cognitive,feldon2007cognitive}.  When structural interference exceeds a moderate threshold  ($I_t^{\text{struct}} > 0.30$), behavioral signals are down-weighted  because observed struggle patterns (e.g., pauses, rapid actions)  increasingly reflect resource depletion rather than informative  strategy shifts \cite{kalyuga2011cognitive}.  This threshold was calibrated to the empirical 30th percentile of  structural load across action-level telemetry ($n=26,188$ drag actions),  a value consistent with the common practice of using percentile-based  splits to demarcate low/high cognitive load conditions in task-complexity  research \cite{paas1994variability}.  Sensitivity analysis confirms that varying the threshold within $\pm 0.05$  yields stable downstream prediction performance (AUC range: $0.829$--$0.850$)  while producing expected shifts in High-Demand Failure prevalence  ($30.2$\%--$51.0$\%; Appendix B, Section 2.3).  This supports the $0.30$ threshold as a robust diagnostic boundary  that balances sensitivity with specificity, grounded in theory  and empirically validated. CDI is initially defined for drag actions and temporally aligned to validation events via forward-fill propagation (Equation 31), ensuring every validation inherits the demand state of the preceding reasoning process. Full calibration procedures and threshold derivation logic are detailed in Equations 32--33 (Appendix A).

\subsection{Quadrant Assignment \& Transition Dynamics}
\label{sec:quadrants_transitions}
Each validation event is assigned to one of four reasoning states using a 2$\times$2 matrix that cross-classifies aligned CDI ($I_v^{\text{filled}}$) and validation outcome: Sound Reasoning (low demand, success), Low-Demand Failure (low demand, failure), Fragile Success (high demand, success), and High-Demand Failure (high demand, failure) (Equation 34). The CDI threshold $\tau = 0.35$ corresponds to the empirical median coordination demand, providing a data-driven split that balances quadrant representation while maintaining theoretical alignment with moderate structural interference. Sensitivity analysis across $\tau \in [0.30, 0.40]$ confirms that key findings (e.g., High-Demand Failure persistence) are robust to threshold selection (Appendix B, Section 2.2).

Transition dynamics capture how reasoning states evolve between consecutive validations. Strategy displacement ($\Delta_{\text{strat}}$) is computed as the Euclidean distance between 5D strategy vectors, while demand shift ($\Delta_{\text{CDI}}$) and error reduction ($\Delta_{\text{hamming}}$) track load and accuracy changes (Equations 35--37). Transitions are classified into four categories using adaptive percentiles (30th/90th of $\Delta_{\text{strat}}$, floored at 0.05) and fixed demand/benefit thresholds (Equations 38--41). Adaptive percentiles account for dataset-specific strategy variability; fixed thresholds correspond to minimally meaningful demand shifts validated against pilot data. Full derivation procedures and robustness checks are provided in Appendix B, Section 2.2.
\begin{itemize}
    \item \textbf{Panic}: high displacement, increasing demand, and error accumulation (Equation 38)
    \item \textbf{Strategic}: moderate displacement, stable/decreasing demand, and meaningful error reduction $\geq 0.5$ (Equation 39)
    \item \textbf{Stable}: minimal strategy shift ($<0.05$) (Equation 40)
    \item \textbf{Adaptive}: all other transitions reflecting context-responsive adjustment (Equation 41)
\end{itemize}

\subsection{Personalization Potential (Variance Decomposition)}
\label{sec:personalization}
To determine whether strategy use reflects stable learner traits or context-dependent states, we compute intraclass correlation coefficients following Shrout \& Fleiss \cite{shrout1979intraclass} for ICC(1) one-way random effects (Equations 42--45). Variance is decomposed into student-level, session-level, and residual components (Equations 46--47). An ICC $< 0.2$ indicates state-dominant variation, where within-student variance across sessions substantially exceeds between-student differences. This statistical property directly informs scaffolding design: low ICC values justify context-aware, real-time adaptation over static learner profiling, as reasoning strategies fluctuate significantly based on exercise structure and immediate coordination demand.

\subsection{Early Warning Prediction Model}
\label{sec:prediction}
To enable proactive intervention, we train a logistic regression model to forecast validation success using pre-validation features. We evaluate a hierarchy of nested models (M0--M3) to isolate component contributions; full results and ablation details are provided in Appendix B, Section 4.1. For the main model, the feature vector $\mathbf{x}_v$ comprises the five concurrent strategy scores, two context variables (actions since last validation, prior quadrant), and two temporal derivatives: Hamming distance trend since the previous validation ($\delta_h$) and strategy stability over the last five validations ($\sigma_S$), computed as the mean standard deviation across all five strategy dimensions (Equations 48--51). Critically, current Hamming distance is explicitly excluded to prevent circularity, ensuring predictive power derives from reasoning patterns rather than direct error counts. 

We deliberately select interpretable logistic regression with $L_2$ regularization over complex black-box architectures. As Rudin \cite{rudin2019stop} argues, high-stakes educational decisions require transparent models whose decision logic can be audited, trusted, and aligned with pedagogical goals. Logistic regression provides coefficient-level interpretability, enabling educators and system designers to trace every prediction back to specific strategy configurations and demand estimates. This choice is further justified by evidence that sequential behavioral patterns reliably predict learner states when properly engineered \cite{fatahi2018empirical}, making non-linear complexity unnecessary for achieving robust discriminative performance.

The model is trained via stratified 5-fold cross-validation with class-weight balancing to address the 38.1\% success rate imbalance. Performance is evaluated using AUC, Average Precision, and accuracy at threshold 0.5. Feature importance is reported via standardized coefficients ($\hat{w}_j / \text{std}(x_j)$), ensuring comparability across heterogeneous feature scales. Threshold optimization for scaffolding triggers (balancing early detection with hint efficiency) and comprehensive robustness checks (generalization, signal stability, calibration) are documented in Appendix B, Sections 4--5 and 6.1. 

\subsection{Scaffolding Policy Integration}
\label{sec:scaffolding_policy}
Diagnostic signals feed directly into a transparent, validation-state decision policy for adaptive scaffolding. While this section outlines the intended mapping from diagnostic states to hint types, the pedagogical efficacy of these mappings has to be empirically validated in a controlled trial. The policy operates on two complementary layers: (1) a \textit{state-based trigger} that evaluates concurrent strategy configurations and coordination demand, and (2) a \textit{trajectory modulator} that incorporates temporal features ($\delta_h$ for error-trend direction, $\sigma_S$ for recent strategy stability) to optimize intervention timing. 

Concretely, when structural interference exceeds $\tau = 0.35$ and Error-Focus drops below baseline, the system identifies a High-Demand Failure state and can trigger facilitative hints that prompt re-evaluation rather than direct correction. The trajectory modulator refines this logic: if $\delta_h$ indicates an accelerating error trend or $\sigma_S$ signals erratic strategy switching, the trigger threshold could be adaptively lowered in future implementations to enable earlier, proactive support. Conversely, stable reasoning with low demand ($\tau < 0.35$, high $S_{\text{focus}}$, low $\sigma_S$) preserves learner autonomy, avoiding hint fatigue and supporting productive struggle. This dual-layer architecture operationalizes stealth assessment principles, inferring latent reasoning states from observable telemetry without interrupting the natural learning flow \cite{shute2011stealth}.

Crucially, all temporal features are inherently interpretable: $\delta_h$ captures recent error-accumulation direction (not current error counts, preserving non-circularity), and $\sigma_S$ reflects short-term strategy consistency. Because both are computed within the same validation-level window as core strategy scores, every trigger remains auditable and traceable to concrete reasoning dynamics. The framework is designed for teacher-AI complementarity rather than full automation; transparent trigger logic enables educators to audit, override, or refine scaffolding decisions based on classroom context and pedagogical priorities \cite{holstein2019codesigning}. The conceptual decision logic is detailed in this section and Section~\ref{sec:prediction}, while threshold optimization results supporting risk-stratified intervention are documented in Appendix B, Section 6.1.

\section{Empirical Results}

\subsection{Why Global Profiling Fails: Heterogeneous Strategy Evolution}
\label{sec:profiling_limitations}
\textit{[Core Finding: State-dominant variation]} To determine whether adaptive scaffolding requires real-time diagnosis or could rely on static learner profiles, we first examined whether students exhibit stable, trait-like reasoning patterns across sessions for each of the 234 students. For each strategy dimension, we computed the per-student slope of strategy score versus validation index (linear regression), classifying trajectories as \emph{increasing} (slope $> +0.01$), \emph{decreasing} (slope $< -0.01$), or \emph{stable} (|$slope$| $\leq 0.01$).

Table~\ref{tab:strategy_evolution} summarizes the distribution of trajectory types. Critically, no strategy shows a dominant directional trend: for each dimension, the majority of students exhibit stable trajectories (62.6--74.5\%), with smaller, heterogeneous splits between increasing and decreasing patterns. This heterogeneity confirms that strategy use is not converging toward a uniform "expert" pattern over time.

\begin{table}[h]
\centering
\caption{Per-student strategy evolution trajectories (N = 234 students)}
\label{tab:strategy_evolution}
\begin{tabular}{lccc}
\toprule
\textbf{Strategy} & \textbf{Increasing (\%)} & \textbf{Stable (\%)} & \textbf{Decreasing (\%)} \\
\midrule
$S_{\text{verb}}$ & 24.2 & 62.6 & 13.2 \\
$S_{\text{chain}}$ & 9.6 & 74.5 & 15.9 \\
$S_{\text{rand}}$ & 14.9 & 72.1 & 13.0 \\
$S_{\text{focus}}$ & 14.6 & 67.6 & 17.8 \\
$S_{\text{strat}}$ & 17.8 & 63.4 & 18.8 \\
\bottomrule
\end{tabular}
\end{table}

Table~\ref{tab:avg_slope} reports the mean slope and standard deviation per strategy. All mean slopes are near-zero (range: $-0.004$ to $+0.010$), with notable inter-student variability (SD $\approx$ 0.031--0.047). This confirms that, on average, students do not systematically increase or decrease any strategy dimension over time; rather, individual trajectories fluctuate based on immediate task demands and coordination demand.

\begin{table}[h]
\centering
\caption{Average strategy evolution slope per dimension (per-student linear regression)}
\label{tab:avg_slope}
\begin{tabular}{lcc}
\toprule
\textbf{Strategy} & \textbf{Mean slope} & \textbf{Std. dev.} \\
\midrule
$S_{\text{verb}}$ & $+0.010$ & 0.035 \\
$S_{\text{chain}}$ & $-0.002$ & 0.031 \\
$S_{\text{rand}}$ & $-0.001$ & 0.034 \\
$S_{\text{focus}}$ & $+0.001$ & 0.047 \\
$S_{\text{strat}}$ & $-0.004$ & 0.046 \\
\bottomrule
\end{tabular}
\end{table}

These findings align with the intraclass correlation results (mean ICC = 0.14, Section~\ref{sec:icc_analysis}): strategy use is state-dominant, not trait-dominant. Consequently, global profiling or longitudinal growth modeling would misrepresent the dynamic, context-dependent nature of grammatical reasoning. Adaptive scaffolding must therefore respond to the student's \emph{current} reasoning configuration rather than historical averages.

\subsection{Strategy Distinctness \& Non-Collinearity}
\label{sec:strategy_distinctness}
A prerequisite for concurrent strategy computation is that dimensions capture theoretically distinct constructs rather than redundant signals. Pairwise Pearson correlations among the five strategy dimensions confirm moderate inter-strategy relationships (max $r = 0.476$ between Randomness and Strategic Commitment), with no pairs exceeding the 0.60 threshold commonly used to indicate problematic multicollinearity. Variance Inflation Factors (VIF) for all strategy features are well below 2.0 (ideal threshold < 5), confirming that multicollinearity does not meaningfully inflate coefficient variance. This empirical distinctness validates the framework's design principle: each dimension quantifies a unique reasoning pattern, justifying their concurrent computation and independent interpretation. $L_2$ regularization (ridge regression) in the prediction model further mitigates any residual multicollinearity, by penalizing large coefficient values, ensuring stable estimates even when predictors exhibit modest correlations \cite{hoerl1970ridge}. Full correlation matrices and robustness checks are documented in Appendix B, Section 6.2.

\subsection{Strategy Transition Dynamics}
\label{sec:transition_dynamics}
\textit{[Core Finding: Adaptive transitions dominate]} Having established that strategies vary by context, our analysis of how students transition between reasoning states across consecutive validations reveals that students continuously adjust their reasoning strategies rather than adhering to fixed approaches. Mean strategy displacement in 5D space is 0.484 (Euclidean), while net CDI change is near-zero (+0.006), indicating balanced coordination demand management across reasoning cycles. Mean Hamming benefit (+0.40 errors reduced per transition) confirms that strategy shifts generally yield measurable accuracy gains.

Transition classification shows that adaptive adjustments dominate (45.9\%), reflecting context-responsive strategy refinement. Stable transitions (38.5\%) indicate periods of consistent reasoning, while strategic shifts (15.3\%) capture deliberate hypothesis exploration. Panic transitions---high-cost, low-benefit thrashing---are rare (0.3\%), suggesting severe unproductive persistence is uncommon at the validation granularity. This pattern aligns with Siegler's \cite{siegler2007cognitive} overlapping waves theory: cognitive variability is not noise but a signature of learning, as students fluidly combine and test multiple reasoning approaches based on immediate feedback and task demands. The predominance of adaptive transitions supports scaffolding policies that preserve moderate strategy exploration while intervening only when configurations become persistently unproductive.

\begin{figure}[h]
\centering
\includegraphics[width=1\textwidth]{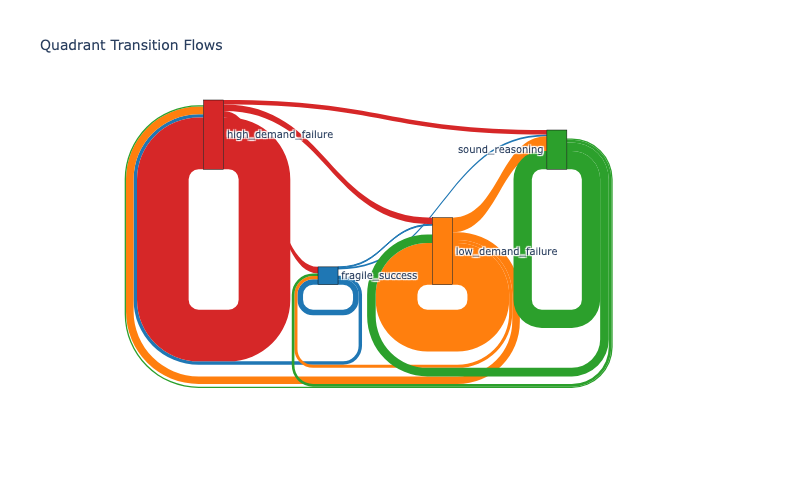}
\caption{\emph{Visual guide:} Vertical bars represent the four reasoning states; flowing bands represent the proportion of students transitioning from validation $t$ to $t+1$. The most prominent visual features are the large, curved self-loops (bands exiting and re-entering the same bar). \emph{Key findings:} The massive red self-loop on High-Demand Failure (74.6\%) and the large orange self-loop on Low-Demand Failure (62.3\%) visually dominate, confirming that students rarely escape unproductive states without intervention. Conversely, the blue Fragile Success state shows a strong outward flow toward the green Sound Reasoning state (52.4\%).}
\label{fig:transition_sankey}
\end{figure}

\subsection{Quadrant Persistence \& State Dynamics}
\label{sec:quadrant_analysis}
\textit{[Core Finding: High persistence in unproductive states]} The critical question for scaffolding design is whether students naturally recover from unproductive states or remain trapped without intervention. Our analysis reveals they largely remain trapped: 74.6\% of High-Demand Failure validations are followed by another High-Demand Failure state, with only 16.0\% escaping to productive states without intervention. Low-Demand Failure states show similar persistence (62.3\% self-loop), while Fragile Success transitions predominantly flow toward Sound Reasoning (52.4\%), indicating successful strategy consolidation when demand is managed.

These transition dynamics are visualized in Figure~\ref{fig:transition_sankey}. The visual dominance of the self-loops (the thick bands looping back onto the High-Demand and Low-Demand bars) immediately illustrates the structural "traps" in student reasoning, highlighting their limited spontaneous recovery from these configurations. Threshold sensitivity analysis confirms robustness: High-Demand Failure persistence ranges 72.8--76.3\% across $\tau \in [0.30, 0.40]$ (Appendix B, Section 2.2), indicating that findings are not artifacts of arbitrary cutoff selection. The quadrant framework thus provides a validated, interpretable lens for distinguishing between students who lack grammatical knowledge (Low-Demand Failure) and those who understand rules but are overwhelmed by coordination demands (High-Demand Failure)---a distinction critical for personalized scaffolding design.

\subsection{Personalization Potential: State-Dominant Strategy Use}
\label{sec:icc_analysis}
Intraclass correlation analysis reveals that strategy use is fundamentally state-dominant rather than trait-like. Mean ICC across five dimensions is 0.139, with individual values ranging 0.092--0.200 (all $< 0.20$). This indicates that a student's reasoning approach varies nearly 7$\times$ more across their own sessions than between different students. Variance decomposition confirms that within-student, across-session variation accounts for the majority of total strategy variance.

This statistical property has direct implications for scaffolding design: low ICC values justify context-aware, real-time adaptation over static learner profiling. Because strategy configurations fluctuate significantly based on exercise structure, immediate coordination demand, and validation history, effective support must respond to the student's current reasoning state rather than historical averages. The framework's validation-level computation (Equations 19--20, Appendix A) operationalizes this principle by anchoring diagnostic signals to the immediate reasoning window.

To further disentangle learner-state variance from task-structure variance, we computed ICC within homogeneous task-complexity bands (2--4, 5--8, 9--13 slots). Results show modestly higher ICC in low-complexity bands but consistently low ICC in medium/high-complexity contexts (overall range: 0.01--0.28; Appendix B, Section 3). This confirms that while task structure partially influences strategy selection in simple exercises, state-dominant variation prevails in complex reasoning tasks, aligning with the framework's focus on real-time, configuration-aware scaffolding.

\subsection{Early Warning Prediction Performance}
\label{sec:prediction_results}

\textit{[Core Finding: Strong prediction without circularity]} The logistic regression model forecasting validation success achieves strong discriminative performance. The strategy-only model (five strategy dimensions + context features, no sequential context) attains AUC = 0.826 (5-fold CV), substantially exceeding typical educational prediction baselines. Adding sequential context (prior quadrant + Hamming trend) yields the full model with AUC = 0.902. For the strategy-only model, bootstrap 95\% CIs [0.818, 0.839] and confirm statistical stability, with Average Precision = 0.714 and accuracy at threshold 0.5 = 0.755. The full model achieves AP = 0.741 and accuracy = 0.826 with bootstrap 95\% CIs [0.894, 0.909]. Both models perform reliably on imbalanced data (38.1\% success rate). Notably, the strategy-only model (AUC = 0.826) outperforms a conventional ITS  baseline using attempt number, time-on-task, prior success rate, and rolling  error count (AUC = 0.780; Appendix B, Section 4.2)---despite containing none  of these standard struggle indicators. This confirms that theoretically-derived  reasoning configurations provide interpretable, actionable diagnostics that  conventional behavioral aggregates cannot replicate.

\begin{figure}[htbp]
\centering
\begin{minipage}{0.48\textwidth}
    \centering
    \textbf{(a) Strategy-Only Model}\\[0.5em]
    \includegraphics[width=\textwidth]{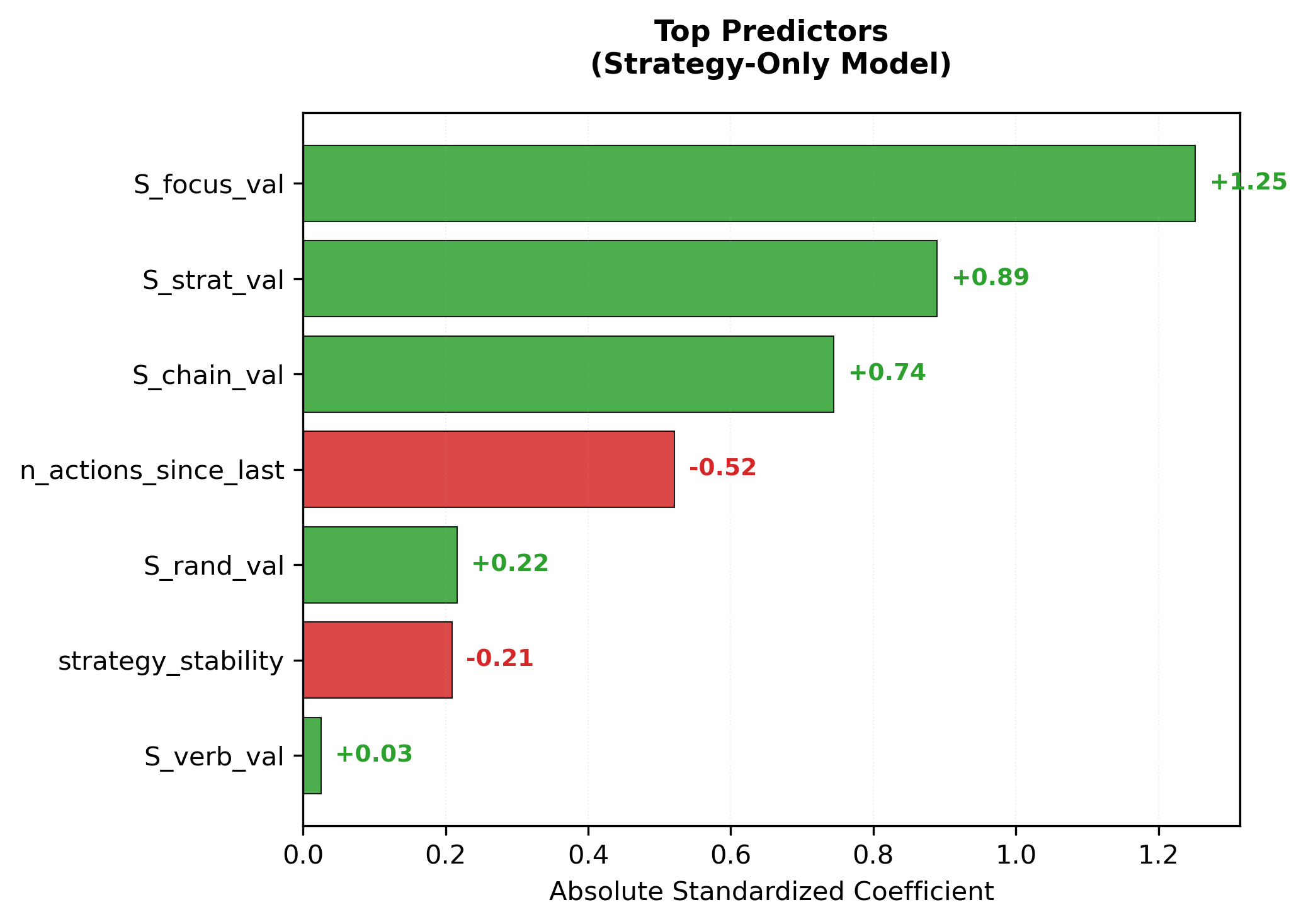}
\end{minipage}
\hfill
\begin{minipage}{0.48\textwidth}
    \centering
    \textbf{(b) Full Sequential Model}\\[0.5em]
    \includegraphics[width=\textwidth]{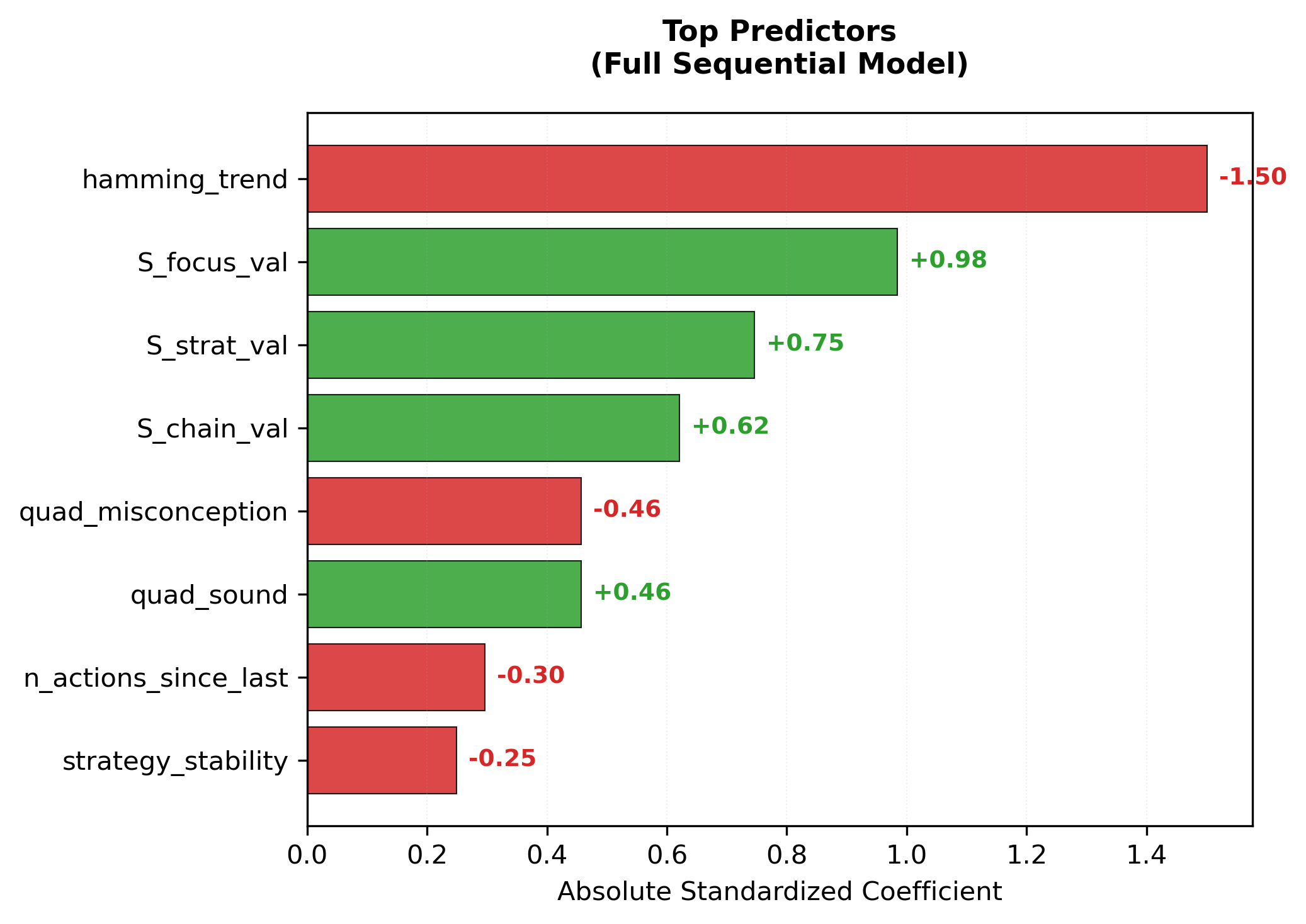}
\end{minipage}
\caption{Top predictors of validation success. Standardized logistic regression coefficients shown; bar length reflects magnitude, sign indicates direction of effect. Error-Focus ($S_{\text{focus}}^{\text{(val)}}$) is the strongest \emph{strategy} predictor in both models; Hamming trend emerges as the overall strongest predictor in the full sequential model.}
\label{fig:feature_importance}
\end{figure}

Feature importance analysis reveals that Error-Focus ($S_{\text{focus}}^{\text{(val)}}$) is the strongest \emph{strategy} predictor in both the strategy-only model ($\beta_{\text{std}} = +1.25$; unstandardized $\beta = +4.79$) and full model (+ sequential context, $\beta_{\text{std}} = +0.98$; unstandardized $\beta = +3.71$; Figure~\ref{fig:feature_importance}), confirming that monitoring whether students systematically target mismatched slots enables reliable detection of struggle before validation failure. In the full model, Hamming trend emerges as the overall strongest predictor ($\beta_{\text{std}} = -1.50$), capturing the rate of error accumulation. While this reflects the incremental value of sequential context, we explicitly tested whether predictive performance was driven primarily by error-accumulation patterns or by the concurrent strategy dimensions. A nested model comparison revealed that the strategy-only configuration (AUC = 0.826) substantially outperforms a Hamming-trend-only baseline (AUC = 0.791), confirming that reasoning configurations are the primary diagnostic signal. Strategy stability emerges as a negative predictor ($\beta_{\text{std}} = -0.21$ strategy-only; $\beta_{\text{std}} = -0.25$ full), indicating that erratic strategy switching signals impending difficulty. Prior quadrant encoding ($\beta_{\text{std}} = -0.46$ for misconception, $+0.46$ for sound) further contributes, aligning with autoregressive expectations that historical outcome informs next-step prediction. Critically, because current Hamming distance is explicitly excluded, both models remain strictly non-circular with respect to direct error derivatives. Full ablation results and model comparisons are provided in Appendix B, Sections 4.4, 6.3, and 6.4.

Notably, Verb-First targeting ($S_{\text{verb}}^{\text{(val)}}$) exhibits near-zero independent predictive power ($\beta \approx -0.02$) despite its theoretical importance in prioritizing high-impact grammatical dependencies. This weak independent signal reflects the fact that verb targeting addresses only a single dependency node, whereas successful reasoning requires coordinating multiple concurrent agreement chains. Consequently, the structural benefits of early verb anchoring are largely subsumed by the broader Chain-Aware progression and Error-Focus dimensions. Rather than serving as a direct predictor of validation success, verb-chain awareness functions as an indirect coordination scaffold---stabilizing the initial reasoning state but contributing little unique predictive variance once global chain management and error-correction precision are accounted for.

To validate that predictive power derives from reasoning configurations rather than sequential outcome history alone, we compared the strategy-only and full (+ sequential context) models. Results show a meaningful performance gain: AUC increases from 0.826 to 0.902 ($\Delta = +0.076$; Appendix B, Section 6.4), confirming that prior outcome and error trends provide useful contextual signal. Crucially, the strategy-only model maintains robust discriminative performance (AUC = 0.826), demonstrating that concurrent strategy configurations and coordination demand estimates contribute substantial, non-redundant predictive value beyond simple prior-success/failure patterns and direct error derivatives. Threshold optimization for scaffolding triggers (balancing early detection with hint efficiency) is documented in Appendix B, Section 6.1.

Figure~\ref{fig:calibration} shows the model's calibration curve: predicted probabilities align closely with actual success rates in the decision-relevant range (0.3--0.7), where scaffolding triggers operate. This confirms that probability estimates are trustworthy for risk-stratified intervention, while extremes (0.0--0.3, 0.7--1.0) exhibit higher variance due to sparse data---a common pattern in imbalanced educational classification.

\begin{figure}[htbp]
\centering
\begin{minipage}{0.48\textwidth}
    \centering
    \textbf{(a) Strategy-Only Model}\\[0.5em]
    \includegraphics[width=\textwidth]{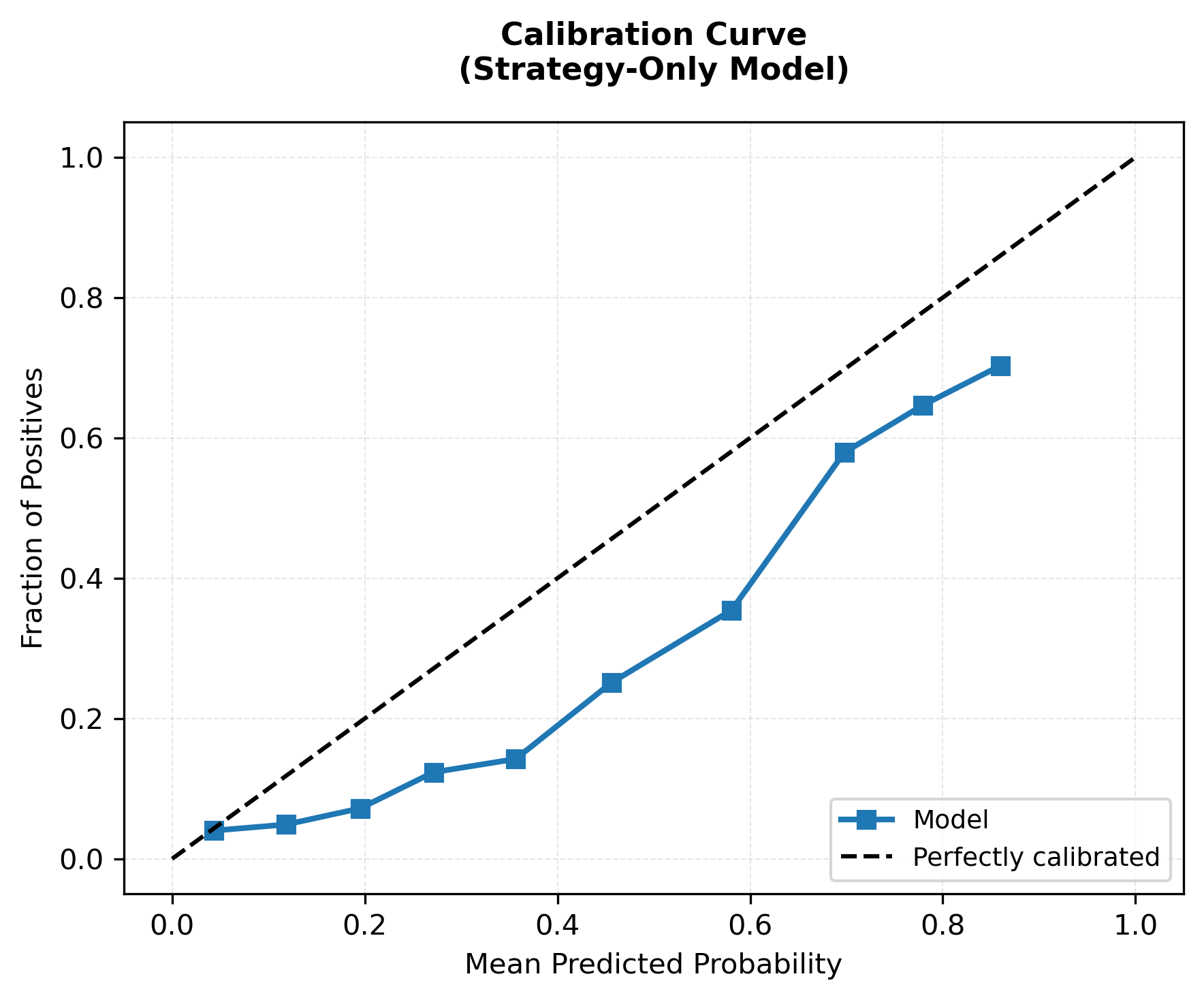}
\end{minipage}
\hfill
\begin{minipage}{0.48\textwidth}
    \centering
    \textbf{(b) Full Sequential Model}\\[0.5em]
    \includegraphics[width=\textwidth]{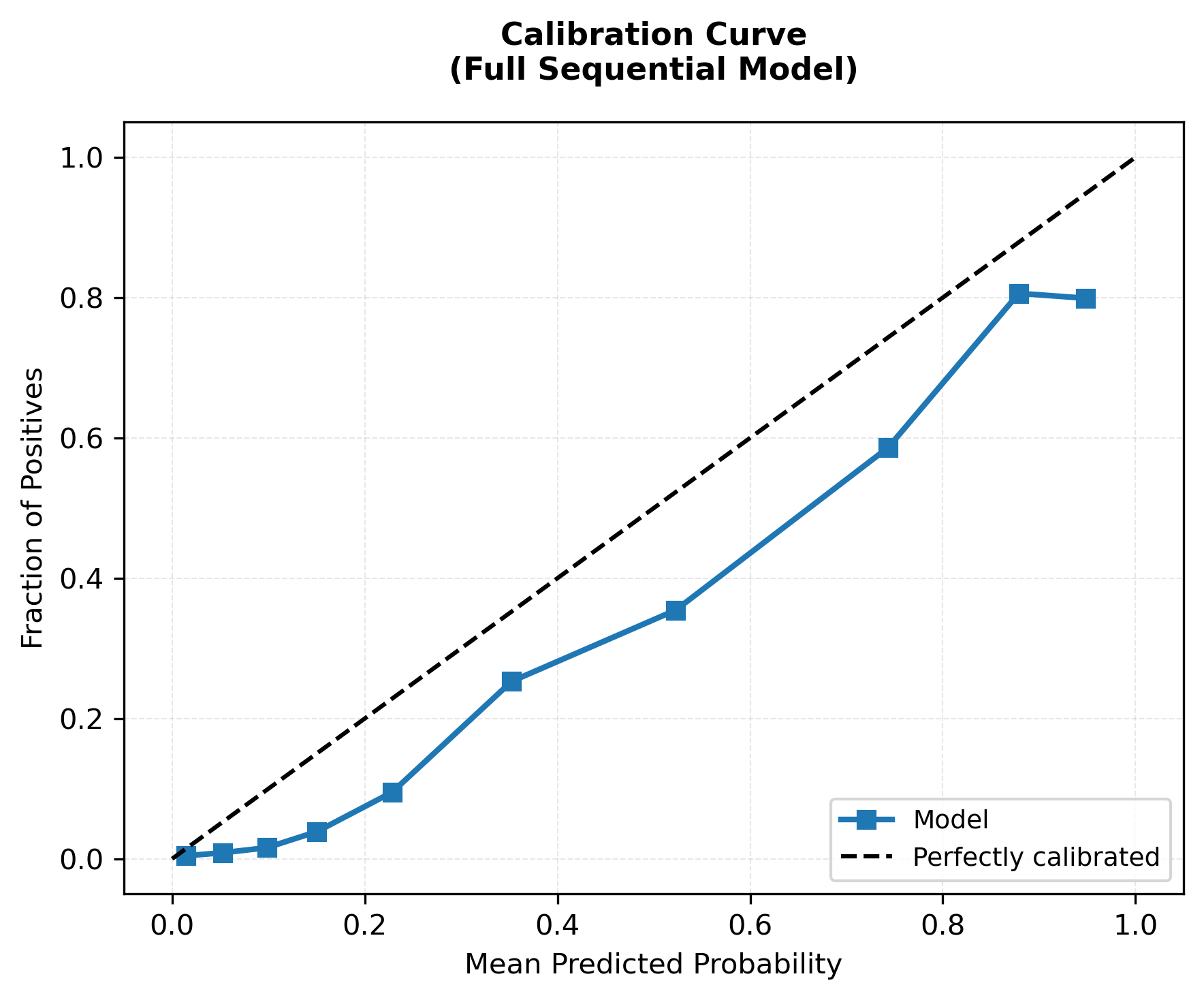}
\end{minipage}
\caption{Model calibration curves. Predicted probabilities (x-axis) vs. actual validation success rates (y-axis) across decile bins. Dashed line indicates perfect calibration. Strong alignment in the 0.3--0.7 range for both models confirms reliable risk estimates for scaffolding triggers.}
\label{fig:calibration}
\end{figure}

This performance substantially exceeds typical ITS prediction baselines, demonstrating that configuration-aware diagnosis provides unique predictive value beyond traditional knowledge-component modeling. The focus on well-defined morphosyntactic patterns---where rules are discrete but application is complex---aligns with meta-analytic evidence that intelligent tutoring is particularly effective in structured domains \cite{steenbergenhu2013meta}. Our framework extends this tradition by modeling not just whether students know a rule, but how they reason about it in real time.

\subsection{Strategy Interactions: Configurational Effects}
\label{sec:interactions}
To test whether strategy effects are purely additive or interact configurationally, we trained a logistic regression model including all 10 pairwise interaction terms among the five strategy dimensions. Table~\ref{tab:all_interactions} reports statistically significant interactions. All 9 significant effects survive Bonferroni correction for 10 comparisons at $\alpha = 0.05$ (corrected threshold $p < 0.005$). The strongest antagonistic interaction occurs between Chain-Aware and Randomness ($\beta = -1.475$, $p_{\text{Bonf}} < 0.001$), indicating that undirected exploration undermines systematic chain completion. Conversely, Chain-Aware combined with Strategic Commitment ($\beta = +1.363$, $p_{\text{Bonf}} < 0.001$) shows strong synergy, suggesting that sustained hypothesis pursuit amplifies the diagnostic value of systematic progression.

\begin{table}[h]
\centering
\small
\caption{Pairwise strategy interactions in early warning model. All reported effects survive Bonferroni correction for 10 comparisons at $\alpha = 0.05$.}
\label{tab:all_interactions}
\begin{tabular}{lccl}
\toprule
\textbf{Interaction} & \textbf{$\beta$} & \textbf{$p$ (raw)} & \textbf{Effect} \\
\midrule
Chain $\times$ Rand & $-1.475$ & $<0.001$ & Antagonistic \\
Chain $\times$ Strat & $+1.363$ & $<0.001$ & Synergistic \\
Rand $\times$ Strat & $+0.930$ & $<0.001$ & Synergistic \\
Focus $\times$ Strat & $+0.291$ & $<0.001$ & Mild synergy \\
Rand $\times$ Focus & $+0.561$ & $<0.001$ & Synergistic \\
Verb $\times$ Chain & $+0.607$ & $<0.001$ & Synergistic \\
Verb $\times$ Rand & $-0.363$ & $<0.001$ & Mild antagonism \\
Verb $\times$ Focus & $+0.207$ & $<0.001$ & Mild synergy \\
Chain $\times$ Focus & $+0.457$ & $<0.001$ & Synergistic \\
\bottomrule
\end{tabular}
\end{table}
\noindent\textbf{Key}: Chain = Chain-Aware, Rand = Randomness, Strat = Strategic Commitment, Focus = Error-Focus, Verb = Verb-First. Full interpretation: antagonistic effects indicate strategy interference; synergistic effects indicate amplification of diagnostic signal.

These configurational effects validate the theoretical premise that grammatical reasoning emerges from dynamic strategy interplay rather than isolated tactics. The predominance of synergistic interactions (7 of 9 significant) suggests that concurrent strategy use generally amplifies diagnostic signal, while antagonistic effects (chain $\times$ randomness, verb $\times$ randomness) indicate that undirected exploration can undermine systematic approaches. Despite these interactions, the additive model achieves strong discriminative performance (held-out AUC = 0.837), confirming that primary strategy effects dominate early warning prediction. This finding supports the current framework's design while motivating future work on interpretable interaction-aware extensions (e.g., constrained generalized additive models) to capture configurational synergies without sacrificing transparency.

\subsection{Robustness \& Generalization}
\label{sec:robustness_summary}

Comprehensive robustness checks confirm that diagnostic signals generalize across students, exercises, and time while remaining stable under realistic telemetry variability. Table~\ref{tab:robustness_main} summarizes key results for both the strategy-only baseline and the full sequential model; full methodological details and per-check parameters are documented in Appendix B, Sections 4, 5, and 7.

\begin{table}[h]
\centering
\caption{Robustness summary: strategy-only vs. full sequential}
\label{tab:robustness_main}
\begin{tabular}{lcc}
\toprule
\textbf{Check} & \textbf{Strategy-Only} & \textbf{Full Sequential} \\
\midrule
CV AUC & 0.826 [0.818, 0.839] & 0.902 [0.894, 0.909] \\
Held-out AUC & 0.837 [0.815, 0.858] & 0.912 [0.897, 0.926] \\
Student-level GroupKFold & 0.823 $\pm$ 0.013 & 0.870 $\pm$ 0.013 \\
Cross-exercise (25\% holdout) & 0.837 & 0.873 \\
Temporal stability (Early/Late) & 0.849 / 0.801 & 0.880 / 0.857 \\
Noise resilience ($\sigma$=0.15) & 0.823 & 0.899 \\
Signal smoothing (window=3) & 0.814 & 0.866 \\
\bottomrule
\end{tabular}
\end{table}

Both models maintain strong generalization and signal stability. The full sequential model shows higher absolute performance across all checks, while the strategy-only model demonstrates that reasoning configurations alone suffice for reliable diagnosis without sequential outcome history.

\subsection{Synthesis: Core Contributions \& Theoretical Positioning}
\label{sec:synthesis}
Table~\ref{tab:synthesis} synthesizes the five core empirical contributions and their scaffolding implications.

\begin{table}
\centering
\caption{Core empirical contributions and their implications for adaptive scaffolding}
\label{tab:synthesis}
\begin{tabular}{p{0.45\textwidth}p{0.45\textwidth}}
\toprule
\textbf{Finding} & \textbf{Implication for Scaffolding} \\
\midrule
Strategy use is state-dominant (mean ICC = 0.14); global profiling fails & Adaptive scaffolding justified; static profiling insufficient \\
\hline
High-Demand Failure persistence = 74.6\% & Students rarely self-correct from high-demand states; targeted intervention needed \\
\hline
Error-Focus predicts success ($\beta = +1.25$); strategies are distinct (max $r = 0.428$) & Monitor error-targeting; trigger hints when focus drops; concurrent dimensions enable nuanced diagnosis \\
\hline
AUC = 0.826 (strategy-only; full + sequential context = 0.902) without circular features; calibration reliable in decision-relevant range (0.3--0.7) & Reasoning patterns alone enable reliable early warning; probabilities trustworthy for scaffolding triggers \\
\hline
Adaptive transitions = 45.9\%; panic transitions = 0.3\% & Preserve moderate exploration; intervene only when configurations become persistently unproductive \\
\bottomrule
\end{tabular}
\end{table}

Collectively, these findings position the framework as a proof-of-concept architecture for real-time, configuration-aware diagnosis in complex reasoning domains. While recent systematic reviews note that language learning remains underrepresented in intelligent tutoring research \cite{mousavinasab2021intelligent}, our work demonstrates that game-based telemetry can yield pedagogically precise diagnostics at scale, bypassing the computational demands of parser-based approaches. The framework's transparency, exercise-awareness, and teacher-trustworthy logic offer a deployable foundation for next-generation scaffolding systems that balance timely intervention with learner autonomy.

\section{Discussion}

\subsection{Theoretical Implications: From Outcome Metrics to Process-Aware Diagnosis}
The empirical findings reposition grammatical reasoning from a binary success/failure paradigm to a dynamic, configuration-aware process. Formative assessment theory emphasizes that effective instruction requires making the gap between current and target performance visible, and providing actionable pathways to close it \cite{black1998assessment}. Traditional ITS implementations often operationalize this gap through error counts or knowledge-component mastery estimates. Our framework advances this tradition by inferring the gap directly from reasoning configurations: high CDI combined with low Error-Focus signals a coordination bottleneck rather than a knowledge deficit, while stable Low-Demand Failure states indicate persistent rule misapplication. This distinction aligns with usage-based acquisition models, which posit that learners extract morphosyntactic regularities through frequency-driven pattern recognition and hypothesis testing \cite{ellis2002frequency}. The variability of the observed strategy (mean ICC = 0.14) confirms that learners do not apply rules uniformly; rather, they fluidly adjust targeting priorities, chain-completion sequences, and error-correction tactics based on immediate structural demands. Notably, while Verb-First targeting shows near-zero independent predictive power ($\beta \approx 0.03$), its theoretical role in prioritizing high-impact grammatical dependencies may manifest indirectly through correlations with Error-Focus and Chain-Aware strategies, suggesting that verb-chain awareness supports coordination primarily as a scaffolding mechanism rather than a direct predictor of success.

Our approach also extends the Intelligent Language Tutoring Systems (ILTS) tradition. Historically, ILTS relied on NLP parsers to analyze open-ended learner input and diagnose agreement, tense, and reference errors with high granularity \cite{heift2007errors}. While parser-based methods yield rich linguistic diagnostics, they are computationally intensive and domain-specific. By contrast, our telemetry-driven pipeline extracts grammatical reasoning patterns at scale from structured interaction logs, bypassing the need for syntactic parsing while preserving diagnostic precision. This demonstrates that game-based action sequences can serve as valid proxies for latent linguistic reasoning, bridging the gap between computational efficiency and pedagogical granularity.

As reported in Section~\ref{sec:interactions}, pairwise interaction analysis confirmed that strategy effects are not purely additive: concurrent use of certain dimensions amplifies or attenuates diagnostic signal. For instance, the antagonistic interaction between Chain-Aware and Randomness ($\beta = -1.475$) suggests that undirected exploration undermines systematic chain completion, while Chain-Aware combined with Strategic Commitment ($\beta = +1.363$) indicates synergistic hypothesis refinement. These configurational effects align with the theoretical premise that grammatical reasoning emerges from dynamic strategy interplay rather than isolated tactics.

\subsection{Scaffolding Policy Design: Transparent, Configuration-Triggered Intervention}
The diagnostic signals can be mapped to a decision logic that prioritizes interpretability and learner autonomy. For example, when structural interference exceeds $\tau = 0.35$ and Error-Focus drops below baseline, the system would identify a High-Demand Failure state and could trigger facilitative hints that prompt re-evaluation rather than direct correction. When prior quadrant classification indicates Low-Demand Failure (low demand, repeated failure), directive hints target specific rule application. Stable reasoning with low demand preserves autonomy, avoiding hint fatigue and supporting productive struggle. This architecture operationalizes Hattie and Timperley's \cite{hattie2007power} feedback model: \emph{Feed Up} (where is the learner in their hypothesis testing?), \emph{Feed Back} (what error or demand pattern emerged?), and \emph{Feed Forward} (what to test next?). For grammatical reasoning, this translates to monitoring Error-Focus ($\beta = +1.25$ in strategy-only model) to detect when students stop targeting mismatches, then selecting hint type based on coordination state. 
Following Shute's \cite{shute2008focus} distinction between directive and facilitative feedback, the framework reserves explicit correction for knowledge gaps and employs metacognitive prompts for coordination bottlenecks. Threshold optimization (Appendix B, Section 6.1) ensures that intervention timing balances early detection with hint efficiency, preventing over-scaffolding while maintaining reliable struggle detection.

It is important to emphasize that this policy architecture is theoretically motivated but not yet empirically validated. Randomized controlled trials are needed to determine whether hints delivered at these diagnostic thresholds improve learning outcomes, reduce unproductive persistence, or accelerate error recovery relative to unassisted practice.

\subsection{Methodological Rigor, Interpretability \& Computational Efficiency}
\label{sec:rigor}

The framework's design prioritizes pedagogical transparency over marginal predictive gains. By selecting logistic regression with $L_2$ regularization and explicitly excluding current Hamming distance, we ensure that every prediction is traceable to specific strategy scores and demand estimates. The ablation analysis confirms that the strategy-only model (no sequential context) achieves AUC = 0.826, while the full model (+ sequential context) attains AUC = 0.902, yet strategy configurations alone sustain strong discriminative performance. This validates the non-circular design principle: predictive power derives from reasoning dynamics, not direct error derivatives. Comprehensive robustness checks (generalization across 42 exercises, temporal stability, noise resilience) confirm that signals are stable under realistic telemetry variability.

From a comparative standpoint, these AUC values place our detector at the upper end of reported performance for real-time disengagement and failure prediction. Liu et al.\ \cite{liu2023struggling} obtained AUC = 0.635 in an open-ended science game, and Karumbaiah et al.\ \cite{karumbaiah2018predicting} reached 0.81 in a physics game. Within grammar ITSs, Deininger et al.\ \cite{deininger2023firsttry} predicted field-level correctness with AUC = 0.68 using XGBoost, and only 0.57 with logistic regression. Even when excluding sequential context and Hamming trend, our strategy-only model (AUC = 0.826) surpasses those benchmarks, and the full model (AUC = 0.902) achieves a level of discriminative power that is rare in educational early warning systems. Crucially, this is accomplished with a transparent logistic regression whose coefficients are directly interpretable, without recourse to post-hoc explanation methods. The comparison suggests that domain structure, when explicitly leveraged through theory-grounded feature engineering, can simultaneously improve predictive accuracy and maintain full pedagogical transparency.

While deep sequence models achieve strong predictive performance in educational data mining, we deliberately selected an interpretable architecture because grammatical reasoning involves pedagogically meaningful patterns (chain completion, error targeting, hypothesis persistence) that should be measured, not only inferred from raw sequences. In structured domains where theoretical constraints are strong, forcing the model to respect pedagogical categories ensures that every scaffolding trigger is traceable to a specific reasoning configuration. This design stance prioritizes human-AI complementarity over automation: educators can audit, override, and refine triggers because the decision logic is transparent, not latent. Furthermore, because the framework outputs low-dimensional, interpretable diagnostic signals rather than opaque latent states, it is designed to interface with arbitrary feedback generation layers---including large language models---while preserving full auditability of the trigger logic.

The deeper rationale for selecting logistic regression is thus epistemic rather than computational. Neural sequence models may achieve marginal predictive gains, but they impose hidden costs that are particularly acute in classroom contexts: opacity under real-time constraints, data hunger in cold-start scenarios, and brittleness under distribution shift when task interfaces evolve. Theory-driven features preserve educator trust and diagnostic actionability in ways that post-hoc explanations of neural architectures cannot replicate.

That said, the pipeline is also computationally efficient enough for real-time classroom deployment. Benchmarking on researcher hardware (Apple M2/8GB RAM) establishes an upper-bound performance profile: the full diagnostic pipeline (feature computation, CDI estimation, and logistic regression inference) completes in $2.2 \pm 0.7$\,ms per validation event (median: $2.0$\,ms, 95th percentile: $3.2$\,ms). Classroom devices (e.g., Intel Celeron/4GB RAM Chromebooks) typically exhibit 3--5$\times$ slower single-threaded performance, yet even under a conservative 5$\times$ slowdown factor, end-to-end processing completes in $<12$\,ms; two orders of magnitude below the typical inter-validation interval ($>2$\,s). Linear computational complexity ($O(K)$ per validation, where $K$ is the number of active features) guarantees predictable scaling across larger cohorts or more complex exercises. Feature computation dominates the pipeline ($1.77$\,ms mean on M2), while CDI estimation ($0.05$\,ms) and model inference ($0.38$\,ms) are negligible, confirming that the framework's diagnostic richness does not compromise deployment feasibility on resource-constrained school hardware.

\subsection{Limitations \& Future Directions}
Several limitations warrant acknowledgment. First, the current validation focuses on French morphosyntax within a single educational context. While the game is deployed in over 20 languages---suggesting architectural reproducibility and cross-linguistic potential---formal validation across typologically distinct languages remains pending. Morphosyntactic complexity, agreement chain density, and orthographic transparency vary significantly across languages; future work will test whether strategy dimensions and CDI thresholds generalize beyond Roman languages. 

Second, the framework currently models cognitive and strategic signals, and the CDI already incorporates behavioral struggle indicators such as pause duration, rapid guessing, slot cycling, and validation pressure. However, it does not explicitly detect or respond to affective states. Control-value theory suggests that prolonged High-Demand Failure states likely co-occur with frustration \cite{pekrun2006control, chitrakar2023frustration}, and affect-aware systems can provide motivational support when such states are detected \cite{dmello2015feeling}. Future iterations will therefore integrate explicit affect-detection channels---for instance, by linking the existing CDI behavioral signals to validated frustration labels or by incorporating multimodal sensors---to deliver scaffolding that balances cognitive challenge with emotional regulation.

Third, the study employs an observational design; while robustness checks confirm predictive validity, causal effects of hint timing and type on learning gains require randomized controlled trials.

Fourth, the ICC = 0.14 finding is among the most theoretically consequential results of this study, as it challenges trait-based assumptions underlying much of adaptive learning. Future work should treat this as a priority. Specifically: (1) Do these low ICCs replicate in other structured reasoning domains, or are they specific to morphosyntax? (2) What is the temporal structure of within-student variation—do strategies fluctuate randomly, or do they follow systematic patterns tied to fatigue, exercise sequencing, or prior success? (3) Under what conditions (e.g., high task complexity) do stable traits re-emerge, as suggested by our complexity-band analysis? A dedicated follow-up exploring the boundary conditions of state-dominant versus trait-dominant variation would have significant implications for how the field conceptualizes learner modeling.

Finally, the linear additive assumption in CDI and logistic regression captures primary effects effectively but may not fully model complex configurational dynamics. Future work will explore targeted pairwise interaction terms and generalized additive models (GAMs) to capture non-linear relationships while preserving the interpretability and computational efficiency required for classroom deployment.

\subsection{A Generalizable Architecture for Structured Reasoning}
\begin{figure}[h]
\centering
\includegraphics[width=1\textwidth]{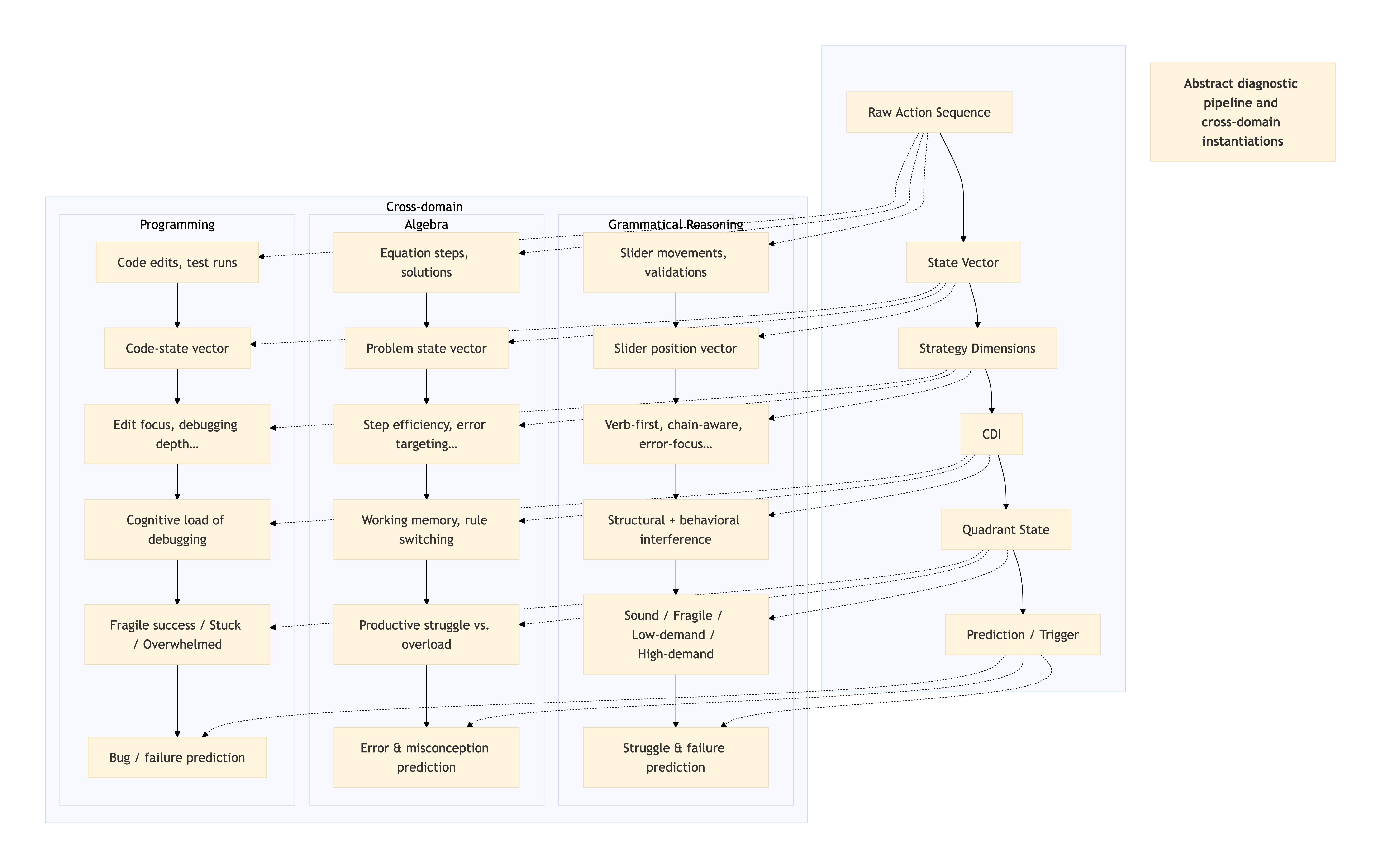}
\caption{Abstract diagnostic pipeline and cross-domain instantiations. Top row: the generalizable architecture from raw actions to interpretable strategy dimensions, CDI estimation, and quadrant-based triggers. Lower panels: instantiations for (a) grammatical reasoning (this paper), (b) algebra equation solving, and (c) programming debugging. In all domains, the framework preserves full interpretability by grounding every diagnostic signal in theoretically defined, auditable reasoning configurations.}
\label{fig:cross_domain}
\end{figure}

While the framework is instantiated here for French morphosyntax, its underlying architecture is domain-agnostic. It provides a transferable blueprint for structured reasoning tasks that produce sequential telemetry over a constrained state space. Such tasks can be diagnosed through the same pipeline: raw actions are mapped to state vectors; strategy dimensions quantify theoretically defined reasoning patterns; a coordination demand index captures structural and behavioral interference; and quadrant states enable transparent, real-time intervention triggers. For example, in algebra equation solving, state vectors represent term assignments, chains represent balancing operations across the left and right sides, and CDI captures interference from competing transformation rules. In programming debugging, states represent variable assignments, chains represent data-flow dependencies, and strategy dimensions distinguish systematic hypothesis testing from random line-tweaking. Figure~\ref{fig:cross_domain} schematizes this abstraction across three domains.

\subsection{Broader Impact on AIED: Toward Hybrid Intelligence in Language Learning}
The framework positions adaptive scaffolding within a hybrid intelligence paradigm, where machine precision complements human pedagogical expertise \cite{dellermann2019hybrid}. AI reliably detects morphosyntactic patterns, quantifies coordination demand, and forecasts struggle probability; teachers interpret creative intent behind ``errors'', adjust hint tone to classroom culture, and override triggers when pedagogical context warrants flexibility. This division of labor preserves the agency of the educator while scaling diagnostic precision beyond human observational capacity.

Methodologically, the pipeline demonstrates that concurrent strategy mining, dynamic demand estimation, and non-circular prediction can be integrated into a single, auditable architecture. While the specific features (verb chains, morphosyntactic slots, slider telemetry) are domain-bound, the underlying approach---extracting latent reasoning configurations from sequential interaction logs and mapping them to interpretable diagnostic states---offers a transferable blueprint for other structured reasoning tasks. Formal cross-domain validation will be required to adapt the framework to mathematics, programming, or scientific inquiry, where constraint navigation and hypothesis testing follow different structural logics. By making reasoning visible, auditable, and actionable, this work advances AIED toward transparent, theory-grounded, and educator-trustworthy design.

\section{Conclusion}

This study introduces a diagnostic framework that transforms opaque grammatical reasoning into observable, actionable signals for adaptive scaffolding. By computing five interpretable strategy dimensions from action sequences, estimating coordination demand in real time, and mapping validations to four distinct reasoning states, the pipeline shifts diagnosis from binary success metrics to dynamic, configuration-aware personalization. Empirical analysis across 2,144 sessions and 42 morphosyntactic exercises demonstrates that strategy use is fundamentally state-dominant (mean ICC = 0.14), with students exhibiting high persistence in unproductive states (74.6\% High-Demand Failure self-loop) and predominantly adaptive strategy transitions (45.9\%). An early warning model forecasting validation success achieves robust discriminative performance (strategy-only AUC = 0.826; full + sequential context AUC = 0.902) while explicitly excluding absolute current error counts to preserve non-circular interpretability.

The framework's strength lies in its commitment to pedagogical transparency: linear logistic regression, ablation-validated feature engineering, and calibrated probability thresholds ensure that every scaffolding trigger is auditable and educator-trustworthy. Computational efficiency ($<12$ ms per validation on classroom hardware) guarantees seamless real-time deployment without disrupting gameplay flow. Extending prior sequence-mining approaches to structured language tasks, this work demonstrates that latent reasoning dynamics can be reliably extracted from fine-grained telemetry, bypassing the computational bottlenecks of parser-based systems while preserving diagnostic precision. Ultimately, by making reasoning visible, contextual, and interpretable, this framework advances the broader AIED vision of artificial intelligence that augments rather than replaces human pedagogy. By providing a diagnostic engine capable of distinguishing how students reason from whether they succeed, it lays the groundwork for scaffolding systems that can respond to reasoning processes in real time.


\section*{Supplementary Materials}
The complete mathematical framework specifying all 52 equations, variable definitions, and computational logic is provided as a separate supplementary file (Appendix A). A comprehensive robustness report documenting all generalization checks, stability analyses, and calibration diagnostics is provided as a second supplementary file (Appendix B). Both appendices are anonymized for double-blind review.

\bibliographystyle{elsarticle-num}
\bibliography{citations}

@article{vanlehn2006behavior,
  author  = {VanLehn, Kurt},
  title   = {The Behavior of Tutoring Systems},
  journal = {International Journal of Artificial Intelligence in Education},
  year    = {2006},
  volume  = {16},
  number  = {3},
  pages   = {227--265}
}

@article{anderson1995cognitive,
  author  = {Anderson, John R. and Corbett, Albert T. and Koedinger, Kenneth R. and Pelletier, Ray},
  title   = {Cognitive Tutors: Lessons Learned},
  journal = {Journal of the Learning Sciences},
  year    = {1995},
  volume  = {4},
  number  = {2},
  pages   = {167--207},
  doi     = {10.1207/s15327809jls0402_2}
}

@incollection{aleven2017instruction,
  author    = {Aleven, Vincent and McLaughlin, Elizabeth A. and Glenn, Robert A. and Koedinger, Kenneth R.},
  title     = {Instruction Based on Adaptive Learning Technologies},
  booktitle = {Handbook of Research on Learning and Instruction},
  editor    = {Mayer, Richard E. and Alexander, Patricia A.},
  edition   = {2nd},
  year      = {2017},
  pages     = {477--518},
  publisher = {Routledge},
  address   = {New York},
  doi       = {10.4324/9781003484776-26}
}

@article{hattie2007power,
  author  = {Hattie, John and Timperley, Helen},
  title   = {The Power of Feedback},
  journal = {Review of Educational Research},
  year    = {2007},
  volume  = {77},
  number  = {1},
  pages   = {81--112},
  doi     = {10.3102/003465430298487}
}

@article{shute2008focus,
  author  = {Shute, Valerie J.},
  title   = {Focus on Formative Feedback},
  journal = {Review of Educational Research},
  year    = {2008},
  volume  = {78},
  number  = {1},
  pages   = {153--189},
  doi     = {10.3102/0034654307313795}
}

@article{sadler1989formative,
  author  = {Sadler, D. Royce},
  title   = {Formative Assessment and the Design of Instructional Systems},
  journal = {Instructional Science},
  year    = {1989},
  volume  = {18},
  number  = {2},
  pages   = {119--144},
  doi     = {10.1007/BF00117714}
}

@article{black1998assessment,
  author  = {Black, Paul and Wiliam, Dylan},
  title   = {Assessment and Classroom Learning},
  journal = {Assessment in Education: Principles, Policy \& Practice},
  year    = {1998},
  volume  = {5},
  number  = {1},
  pages   = {7--74},
  doi     = {10.1080/0969595980050102}
}

@inproceedings{kehrer2013does,
  author    = {Kehrer, Peter and Kelly, Kristen and Heffernan, Neil},
  title     = {Does Immediate Feedback While Doing Homework Improve Learning?},
  booktitle = {Proceedings of the 26th International Florida Artificial Intelligence Research Society Conference (FLAIRS)},
  year      = {2013},
  pages     = {542--545},
  publisher = {AAAI Press},
  address   = {Menlo Park, CA},
  url       = {https://www.aaai.org/ocs/index.php/FLAIRS/FLAIRS13/paper/view/5938}
}

@incollection{shute2011stealth,
  author    = {Shute, Valerie J.},
  title     = {Stealth Assessment in Computer-Based Games to Support Learning},
  booktitle = {Computer Games and Instruction},
  editor    = {Tobias, Sigmund and Fletcher, J. D.},
  year      = {2011},
  pages     = {503--524},
  publisher = {Emerald Publishing Limited},
  doi       = {10.1108/978-1-61735-410-620251025}
}

@article{snow2015agency,
  author  = {Snow, Erica L. and Allen, Laura K. and Jacovina, Matthew E. and McNamara, Danielle S.},
  title   = {Does Agency Matter? Exploring the Impact of Controlled Behaviors within a Game-Based Environment},
  journal = {Computers \& Education},
  year    = {2015},
  volume  = {82},
  pages   = {378--392},
  doi     = {10.1016/j.compedu.2014.12.011}
}

@inproceedings{harpstead2015using,
  author    = {Harpstead, Erik and Aleven, Vincent},
  title     = {Using Empirical Learning Curve Analysis to Inform Design in an Educational Game},
  booktitle = {Proceedings of the 2015 Annual Symposium on Computer-Human Interaction in Play},
  series    = {CHI PLAY '15},
  year      = {2015},
  pages     = {197--207},
  publisher = {Association for Computing Machinery},
  address   = {New York, NY, USA},
  doi       = {10.1145/2793107.2793128}
}

@article{lu2025from,
  author    = {Lu, Wenyi and Griffin, Joseph and Laffey, James and Sadler, Troy D. and Goggins, Sean},
  title     = {From In-Game Behaviors to Learning Gains: Constructing Bayesian Networks for Stealth Assessment},
  journal   = {Proceedings of the ACM on Human-Computer Interaction},
  year      = {2025},
  volume    = {9},
  number    = {6},
  articleno = {GAMES020},
  pages     = {1--40},
  month     = oct,
  publisher = {Association for Computing Machinery},
  address   = {New York, NY, USA},
  doi       = {10.1145/3748615}
}

@book{shute2013stealth,
  author    = {Shute, Valerie J. and Ventura, Matthew},
  title     = {Stealth Assessment: Measuring and Supporting Learning in Video Games},
  year      = {2013},
  publisher = {MIT Press},
  address   = {Cambridge, MA},
  doi       = {10.7551/mitpress/9589.001.0001}
}

@article{flavell1979metacognition,
  author  = {Flavell, John H.},
  title   = {Metacognition and Cognitive Monitoring: A New Area of Cognitive--Developmental Inquiry},
  journal = {American Psychologist},
  year    = {1979},
  volume  = {34},
  number  = {10},
  pages   = {906--911},
  doi     = {10.1037/0003-066X.34.10.906}
}

@incollection{winne1998studying,
  author    = {Winne, Philip H. and Hadwin, Allyson F.},
  title     = {Studying as Self-Regulated Learning},
  booktitle = {Metacognition in Educational Theory and Practice},
  editor    = {Hacker, Douglas J. and Dunlosky, John and Graesser, Arthur C.},
  year      = {1998},
  pages     = {277--304},
  publisher = {Lawrence Erlbaum Associates},
  address   = {Mahwah, NJ}
}

@article{holstein2019codesigning,
  author  = {Holstein, Kenneth and McLaren, Bruce M. and Aleven, Vincent},
  title   = {Co-designing a Real-Time Classroom Orchestration Tool to Support Teacher--AI Complementarity},
  journal = {Journal of Learning Analytics},
  year    = {2019},
  volume  = {6},
  number  = {2},
  pages   = {27--52},
  doi     = {10.18608/jla.2019.62.3}
}

@article{dellermann2019hybrid,
  author  = {Dellermann, Dominik and Ebel, Philipp and S{\"o}llner, Matthias and Leimeister, Jan Marco},
  title   = {Hybrid Intelligence},
  journal = {Business \& Information Systems Engineering},
  year    = {2019},
  volume  = {61},
  number  = {5},
  pages   = {637--643},
  doi     = {10.1007/s12599-019-00595-2}
}

@article{roll2016evolution,
  author  = {Roll, Ido and Wylie, Rebecca},
  title   = {Evolution and Revolution in Artificial Intelligence in Education},
  journal = {International Journal of Artificial Intelligence in Education},
  year    = {2016},
  volume  = {26},
  number  = {2},
  pages   = {582--599},
  doi     = {10.1007/s40593-016-0110-3}
}

@article{rudin2019stop,
  author  = {Rudin, Cynthia},
  title   = {Stop Explaining Black Box Machine Learning Models for High Stakes Decisions and Use Interpretable Models Instead},
  journal = {Nature Machine Intelligence},
  year    = {2019},
  volume  = {1},
  number  = {5},
  pages   = {206--215},
  doi     = {10.1038/s42256-019-0048-x}
}

@book{molnar2020interpretable,
  author    = {Molnar, Christoph},
  title     = {Interpretable Machine Learning: A Guide for Making Black Box Models Explainable},
  year      = {2020},
  publisher = {Lulu.com},
  url       = {https://christophm.github.io/interpretable-ml-book/}
}

@article{carvalho2019machine,
  author  = {Carvalho, Diogo V. and Pereira, Eduardo M. and Cardoso, Jaime S.},
  title   = {Machine Learning Interpretability: A Survey on Methods and Metrics},
  journal = {Electronics},
  year    = {2019},
  volume  = {8},
  number  = {8},
  pages   = {832},
  doi     = {10.3390/electronics8080832}
}

@article{ellis2002frequency,
  author  = {Ellis, Nick C.},
  title   = {Frequency Effects in Language Processing: A Review with Implications for Theories of Implicit and Explicit Language Acquisition},
  journal = {Studies in Second Language Acquisition},
  year    = {2002},
  volume  = {24},
  number  = {2},
  pages   = {143--188},
  doi     = {10.1017/S0272263102002024}
}

@book{dekeyser2007practice,
  author    = {DeKeyser, Robert M.},
  title     = {Practice in a Second Language: Perspectives from Applied Linguistics and Cognitive Psychology},
  year      = {2007},
  publisher = {Cambridge University Press},
  address   = {Cambridge, UK},
  doi       = {10.1017/CBO9780511667275}
}

@book{heift2007errors,
  author    = {Heift, Trude and Schulze, Mathias},
  title     = {Errors and Intelligence in Computer-Assisted Language Learning: Parsers and Pedagogues},
  year      = {2007},
  publisher = {Routledge},
  address   = {New York}
}

@article{ranalli2018automated,
  author  = {Ranalli, Jim},
  title   = {Automated Written Corrective Feedback: How Well Can Students Make Use of It?},
  journal = {Computer Assisted Language Learning},
  year    = {2018},
  volume  = {31},
  number  = {7},
  pages   = {653--674},
  doi     = {10.1080/09588221.2018.1428994}
}

@article{ravid2002developing,
  author  = {Ravid, Dorit and Tolchinsky, Liliana},
  title   = {Developing Linguistic Literacy: A Comprehensive Model},
  journal = {Journal of Child Language},
  year    = {2002},
  volume  = {29},
  number  = {2},
  pages   = {417--447},
  doi     = {10.1017/S0305000902005111}
}

@book{karmiloffsmith1981functional,
  author    = {Karmiloff-Smith, Annette},
  title     = {A Functional Approach to Child Language: A Study of Determiners and Reference},
  year      = {1979},
  publisher = {Cambridge University Press},
  address   = {Cambridge, UK}
}

@book{brissaud2018etude,
  author    = {Brissaud, Catherine and Fayol, Michel},
  title     = {{\'E}tude de la langue et production d'{\'e}crits},
  year      = {2018},
  publisher = {CNESCO},
  note      = {Rapport scientifique, conf{\'e}rence de consensus {\'E}crire et r{\'e}diger}
}

@article{pekrun2006control,
  author  = {Pekrun, Reinhard},
  title   = {The Control-Value Theory of Achievement Emotions: Assumptions, Corollaries, and Implications for Educational Research and Practice},
  journal = {Educational Psychology Review},
  year    = {2006},
  volume  = {18},
  number  = {4},
  pages   = {315--341},
  doi     = {10.1007/s10648-006-9029-9}
}

@incollection{dmello2015feeling,
  author    = {D'Mello, Sidney K. and Graesser, Arthur C.},
  title     = {Feeling, Thinking, and Computing with Affect-Aware Learning Technologies},
  booktitle = {The Oxford Handbook of Affective Computing},
  editor    = {Calvo, Rafael A. and D'Mello, Sidney and Gratch, Jonathan and Kappas, Arvid},
  year      = {2015},
  pages     = {419--434},
  publisher = {Oxford University Press},
  address   = {New York},
  doi       = {10.1093/oxfordhb/9780199942237.013.032}
}

@article{chitrakar2023frustration,
  author  = {Chitrakar, Nandita and P.M., Rajesh},
  title   = {Frustration and Its Influences on Student Motivation and Academic Performance},
  journal = {International Journal of Scientific Research in Modern Science and Technology},
  year    = {2023},
  volume  = {2},
  number  = {11},
  pages   = {1--9},
  doi     = {10.59828/ijsrmst.v2i11.158}
}

@article{kulik2016effectiveness,
  author  = {Kulik, James A. and Fletcher, J. D.},
  title   = {Effectiveness of Intelligent Tutoring Systems: A Meta-Analytic Review},
  journal = {Review of Educational Research},
  year    = {2016},
  volume  = {86},
  number  = {1},
  pages   = {42--78},
  doi     = {10.3102/0034654315581420}
}

@article{steenbergenhu2013meta,
  author  = {Steenbergen-Hu, Saiying and Cooper, Harris},
  title   = {A Meta-Analysis of the Effectiveness of Intelligent Tutoring Systems on {K--12} Students' Mathematical Learning},
  journal = {Journal of Educational Psychology},
  year    = {2013},
  volume  = {105},
  number  = {4},
  pages   = {970--987},
  doi     = {10.1037/a0032447}
}

@article{mousavinasab2021intelligent,
  author  = {Mousavinasab, Elham and Zarifsanaiey, Nahid and Niakan Kalhori, Sharareh R. and Rakhshan, Mahnaz and Keikha, Leila and Ghazi Saeedi, Maryam},
  title   = {Intelligent Tutoring Systems: A Systematic Review of Characteristics, Applications, and Evaluation Methods},
  journal = {Interactive Learning Environments},
  year    = {2018},
  volume  = {29},
  number  = {1},
  pages   = {142--163},
  doi     = {10.1080/10494820.2018.1558257}
}

@article{siegler2007cognitive,
  author  = {Siegler, Robert S.},
  title   = {Cognitive Variability},
  journal = {Developmental Science},
  year    = {2007},
  volume  = {10},
  number  = {1},
  pages   = {104--109},
  doi     = {10.1111/j.1467-7687.2007.00571.x}
}

@article{bakeryacef2009state,
  author  = {Baker, Ryan S. J. d. and Yacef, Kalina},
  title   = {The State of Educational Data Mining in 2009: A Review and Future Visions},
  journal = {Journal of Educational Data Mining},
  year    = {2009},
  volume  = {1},
  number  = {1},
  pages   = {3--17},
  url     = {https://jedm.educationaldatamining.org/index.php/JEDM/article/view/25}
}

@article{bannert2014process,
  author  = {Bannert, Maria and Reimann, Peter and Sonnenberg, Christoph},
  title   = {Process Mining Techniques for Analysing Patterns and Strategies in Students' Self-Regulated Learning},
  journal = {Metacognition and Learning},
  year    = {2014},
  volume  = {9},
  number  = {2},
  pages   = {161--185},
  doi     = {10.1007/s11409-013-9107-6}
}

@article{fatahi2018empirical,
  author  = {Fatahi, Somayeh and Shabanali-Fami, Faezeh and Moradi, Hadi},
  title   = {An Empirical Study of Using Sequential Behavior Pattern Mining Approach to Predict Learning Styles},
  journal = {Education and Information Technologies},
  year    = {2018},
  volume  = {23},
  number  = {4},
  pages   = {1427--1445},
  doi     = {10.1007/s10639-017-9667-1}
}

@inproceedings{bechet2012discovering,
  author    = {B{\'e}chet, Nicolas and Cellier, Peggy and Charnois, Thierry and Cr{\'e}milleux, Bruno},
  title     = {Discovering Linguistic Patterns Using Sequence Mining},
  booktitle = {Proceedings of the 13th International Conference on Intelligent Text Processing and Computational Linguistics (CICLing)},
  year      = {2012},
  pages     = {154--165},
  publisher = {Springer},
  address   = {Berlin, Heidelberg},
  doi       = {10.1007/978-3-642-28604-9_13}
}

@article{hamming1950error,
  author  = {Hamming, Richard W.},
  title   = {Error Detecting and Error Correcting Codes},
  journal = {The Bell System Technical Journal},
  year    = {1950},
  volume  = {29},
  number  = {2},
  pages   = {147--160},
  doi     = {10.1002/j.1538-7305.1950.tb00463.x}
}

@article{shrout1979intraclass,
  author  = {Shrout, Patrick E. and Fleiss, Joseph L.},
  title   = {Intraclass Correlations: Uses in Assessing Rater Reliability},
  journal = {Psychological Bulletin},
  year    = {1979},
  volume  = {86},
  number  = {2},
  pages   = {420--428},
  doi     = {10.1037/0033-2909.86.2.420}
}

@article{dent2016relation,
  author  = {Dent, Allison L. and Koenka, Anna C.},
  title   = {The Relation Between Self-Regulated Learning and Academic Achievement Across Childhood and Adolescence: A Meta-Analysis},
  journal = {Educational Psychology Review},
  year    = {2016},
  volume  = {28},
  number  = {3},
  pages   = {425--474},
  doi     = {10.1007/s10648-015-9320-8}
}

@inproceedings{geoffre2026modeling,
  author    = {Geoffre, Thierry and Geoffre, Trystan},
  title     = {Modeling Grammatical Hypothesis Testing in Young Learners: A Sequence-Based Learning Analytics Study of Morphosyntactic Reasoning in an Interactive Game},
  booktitle = {Proceedings of the 16th International Learning Analytics and Knowledge Conference},
  series    = {LAK '26},
  year      = {2026},
  pages     = {1--10},
  publisher = {Association for Computing Machinery},
  address   = {New York, NY, USA},
  doi       = {10.1145/3785022.3785126}
}

@article{sweller1994cognitive,
  title={Cognitive load theory, learning difficulty, and instructional design},
  author={Sweller, John},
  journal={Learning and Instruction},
  volume={4},
  number={4},
  pages={295--312},
  year={1994},
  publisher={Elsevier},
  doi={10.1016/0959-4752(94)90003-5}
}

@inproceedings{liu2023struggling,
  title={Struggling to Detect Struggle in Students Playing a Science Exploration Game},
  author={Liu, Xiner and Slater, Stefan and Andres, Juliana Ma. Alexandra L. and Swanson, Luke and Scianna, Jennifer and Gagnon, David and Baker, Ryan S.},
  booktitle={Companion Proceedings of the Annual Symposium on Computer-Human Interaction in Play (CHI PLAY Companion '23)},
  year={2023},
  doi={10.1145/3573382.3616080}
}

@inproceedings{karumbaiah2018predicting,
  title={Predicting Quitting in Students Playing a Learning Game},
  author={Karumbaiah, Shamya and Baker, Ryan S. and Shute, Valerie},
  booktitle={Proceedings of the 11th International Conference on Educational Data Mining (EDM 2018)},
  year={2018},
  pages={167--176}
}

@inbook{deininger2023firsttry,
  author={Deininger, Hannah and Lavelle-Hill, Rosa and Parrisius, Cora and Pieronczyk, Ines and Colling, Leona and Meurers, Detmar and Trautwein, Ulrich and Nagengast, Benjamin and Kasneci, Gjergji},
  title={Can You Solve This on the First Try? -- Understanding Exercise Field Performance in an Intelligent Tutoring System},
  booktitle={Artificial Intelligence in Education. AIED 2023. Lecture Notes in Computer Science, vol 13916},
  year={2023},
  pages={565--576},
  publisher={Springer},
  doi={10.1007/978-3-031-36272-9_46}
}

@inproceedings{deininger2025who,
  title={Who Did What to Succeed? Individual Differences in Which Learning Behaviors Are Linked to Achievement},
  author={Deininger, Hannah and Parrisius, Cora and Lavelle-Hill, Rosa and Meurers, Detmar and Trautwein, Ulrich and Nagengast, Benjamin and Kasneci, Gjergji},
  booktitle={Proceedings of the 15th International Learning Analytics and Knowledge Conference (LAK 2025)},
  year={2025},
  doi={10.1145/3706468.3706571}
}

@article{sweller2019cognitive,
  title    = {Cognitive architecture and instructional design: 20 years later},
  author   = {Sweller, John and van Merri{\"e}nboer, Jeroen J. G. and Paas, Fred},
  journal  = {Educational Psychology Review},
  volume   = {31},
  pages    = {261--292},
  year     = {2019},
  doi      = {10.1007/s10648-019-09465-5}
}

@article{feldon2007cognitive,
  title   = {Cognitive load and classroom teaching: The double-edged sword of automaticity},
  author  = {Feldon, David F.},
  journal = {Educational Psychologist},
  volume  = {42},
  number  = {3},
  pages   = {123--137},
  year    = {2007},
  doi     = {10.1080/00461520701416173}
}

@article{kalyuga2011cognitive,
  title   = {Cognitive load theory: How many types of load does it really need?},
  author  = {Kalyuga, Slava},
  journal = {Educational Psychology Review},
  volume  = {23},
  number  = {1},
  pages   = {1--19},
  year    = {2011},
  doi     = {10.1007/s10648-010-9150-7}
}

@article{paas1994variability,
  title   = {Variability of worked examples and transfer of geometrical problem-solving skills: A cognitive-load approach},
  author  = {Paas, Fred and van Merri{\"e}nboer, Jeroen J. G.},
  journal = {Journal of Educational Psychology},
  volume  = {86},
  number  = {1},
  pages   = {122--133},
  year    = {1994},
  doi     = {10.1037/0022-0663.86.1.122}
}

@article{hoerl1970ridge,
  title={Ridge regression: Biased estimation for nonorthogonal problems},
  author={Hoerl, Arthur E and Kennard, Robert W},
  journal={Technometrics},
  volume={12},
  number={1},
  pages={55--67},
  year={1970},
  publisher={Taylor \& Francis}
}

@article{rittlejohnson1999learning,
  author = {Rittle-Johnson, Bethany and Siegler, Robert S.},
  title = {Learning to Spell: Variability, Choice, and Change in Children's Strategy Use},
  journal = {Child Development},
  year = {1999},
  volume = {70},
  number = {2},
  pages = {336--357},
  doi = {10.1111/1467-8624.00025}
}

\end{document}